\documentclass[twoside,twocolumn,10pt]{article}
\usepackage{wscg}           % includes a number of other packages (e.g., myalgorithm)
\RequirePackage{ifpdf}
\ifpdf
 \RequirePackage[pdftex]{graphicx}
 \RequirePackage[pdftex]{color}
\else
 \RequirePackage[dvips,draft]{graphicx}
 \RequirePackage[dvips]{color}
\fi
\usepackage{booktabs}      % typographically much better
\usepackage{multirow}
\usepackage{rotating}
\usepackage{tabularray}
\usepackage{subcaption}
\usepackage[switch]{lineno}

\usepackage{nopageno}       % no page numbers at all; uncomment for final version

\title{Fast Training Data Acquisition for Object Detection and Segmentation using Black Screen Luminance Keying}

\author{
\parbox{0.25\textwidth}{\centering
Thomas P\"ollabauer
}
\hspace{0.05\textwidth}
\parbox{0.25\textwidth}{\centering
Volker Knauthe
}
\hspace{0.05\textwidth}
\parbox{0.25\textwidth}{\centering
Andr\'e Boller
}\\
\vspace{0.1cm}
\hspace{0.05\textwidth}\\
\parbox{0.25\textwidth}{\centering
Arjan Kuijper
}
\hspace{0.05\textwidth}
\parbox{0.25\textwidth}{\centering
Dieter W. Fellner
}\\
\vspace{0.1cm}
\parbox{0.75\textwidth}{\centering
Fraunhofer IGD, Fraunhoferstrasse 5, 64283, Darmstadt, Germany\\
TU Darmstadt, Karolinenplatz 5, 64289, Darmstadt, Germany\\
\vspace{0.1cm}
thomas.poellabauer@igd.fraunhofer.de
}
}

\usepackage{url}
\makeatletter
\def\Uslash{\mathbin{\mathchar`\/}\@ifnextchar{/}{\kern-.15em}{}}
\g@addto@macro\UrlSpecials{\do \/ {\Uslash}}
\def\Ucolon{\mathbin{\mathchar`:}\@ifnextchar{/}{\kern-.1em}{}}
\g@addto@macro\UrlSpecials{\do : {\Ucolon}}
\makeatother

\begin{document}

\twocolumn[{\csname @twocolumnfalse\endcsname

\maketitle  % full width title

% * State the problem 
% * Say why it’s an interesting problem
% * Say what your solution achieves
% * Say what follows from your solution 

\begin{abstract}
\noindent
Deep Neural Networks (DNNs) require large amounts of annotated training data for a good performance. Often this data is generated using manual labeling (error-prone and time-consuming) or rendering (requiring geometry and material information). Both approaches make it difficult or uneconomic to apply them to many small-scale applications. A fast and straightforward approach of acquiring the necessary training data would allow the adoption of deep learning to even the smallest of applications. Chroma keying is the process of replacing a color (usually blue or green) with another background. Instead of chroma keying, we propose luminance keying for fast and straightforward training image acquisition. We deploy a black screen with high light absorption (99.99\%) to record roughly 1-minute long videos of our target objects, circumventing typical problems of chroma keying, such as color bleeding or color overlap between background color and object color. Next we automatically mask our objects using simple brightness thresholding, saving the need for manual annotation. Finally, we automatically place the objects on random backgrounds and train a 2D object detector. We do extensive evaluation of the performance on the widely-used YCB-V object set and compare favourably to other conventional techniques such as rendering, without needing 3D meshes, materials or any other information of our target objects and in a fraction of the time needed for other approaches. Our work demonstrates highly accurate training data acquisition allowing to start training state-of-the-art networks within minutes.

\end{abstract}

\subsection*{Keywords}
Machine Learning, Object Detection, Object Segmentation, Deep Neural Networks.

\vspace*{1.0\baselineskip}
}]

\section{Introduction}
\label{sec:intro}
Modern machine learning (ML) is dominated by deep neural networks (DNNs). Training DNNs to state-of-the-art performance levels tends to require large amounts of training data to perform well. In many cases the lack of annotated data prevents the use of these networks. In the case of object segmentation and object detection a common approach - aside of costly manual labeling - is to resort to rendering to generate the required training images. To do so 3D meshes, textures, and material properties of the target objects are required, which is another barrier for many applications. In contrast, we only require the availability of the objects in question, a camera, a sufficiently large piece of special black cloth, and some lights. We completely circumvent many of the problems with traditional chroma keying such as color bleeding, same foreground/background color and similar, by utilizing a very low reflectance cloth and demonstrate its applicability in a fast, low-cost, high-quality setup, comparing favorably to much more complex data generation regimen. \\

Our contributions are the following:
\begin{itemize}
    \item We propose a straightforward, easy to use setup to record high-quality training datasets for object segmentation and object detection.
    \item We present extensive evaluation and show the performance of our approach in comparison with other conventional data generation methods that require more information, such as meshes, textures, and materials, and/or much more processing time. To make our results more meaningful for the research community, we do all our evaluation on the common YCB-V dataset.
    \item We provide code, which automatically converts the recordings to datasets in COCO format for use with segmentation and 2D object detection algorithms, as well as our black screen recordings of the YCB-V objects.
\end{itemize}

\begin{figure*}[htbp]
    \centering
    \includegraphics[width=\textwidth]{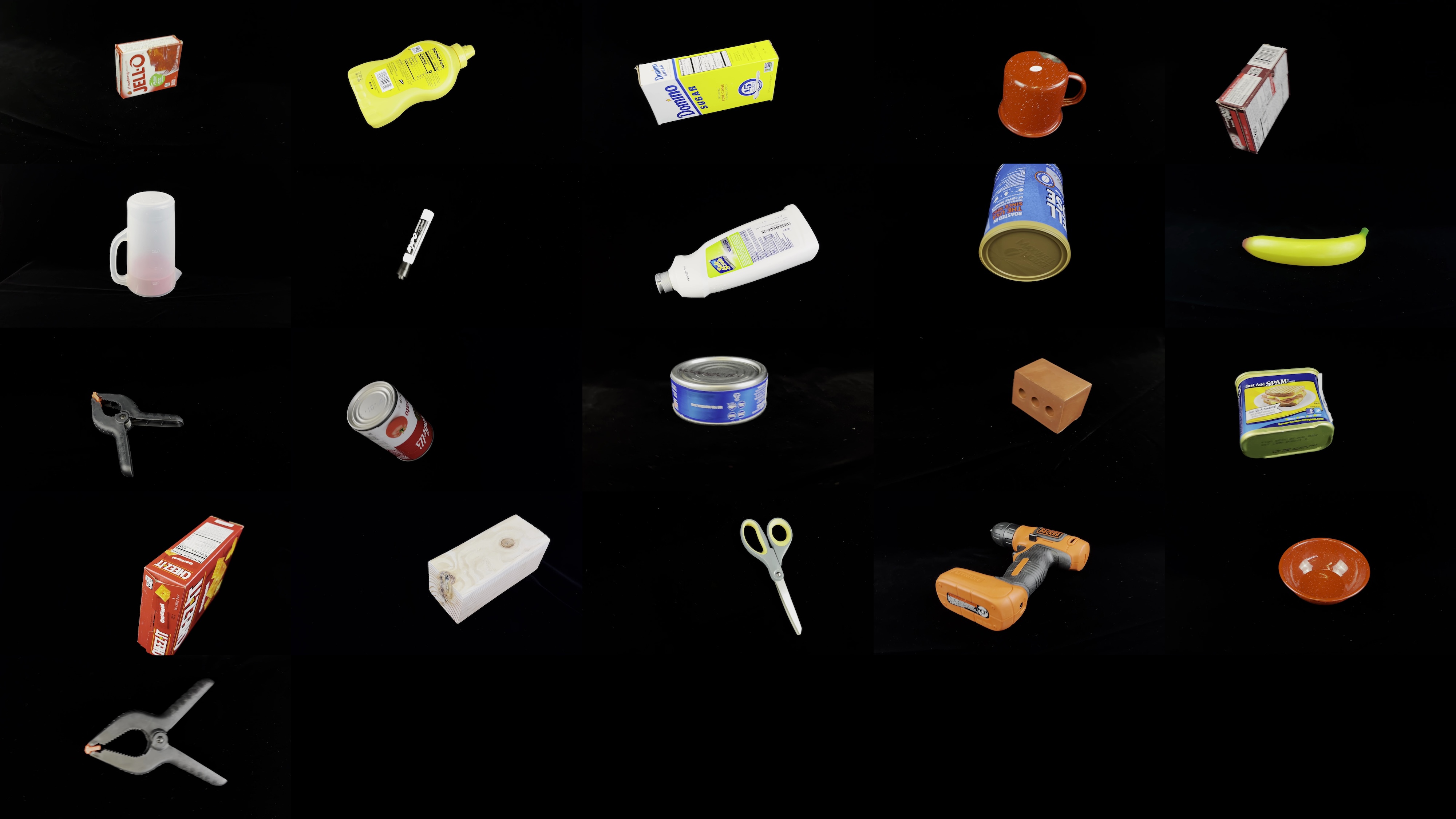}
    \caption{A qualitative sample of all 21 YCB-V objects, that were recorded with a handheld smartphone and our proposed black background. It can be seen that the objects are well silhouetted against the background and can therefore be segmented in an easy way and many typical chroma key-associated problems are circumvented.}
    \label{fig:ycbv_objects}
\end{figure*}

\section{Related Work}
\subsection{Data Generation for ML}
Insufficient training data is a well documented problem in Machine Learning. In the domain of Computer Vision practitioners soon adopted rendering for data generation. Rendering has many suitable attributes, such as perfect ground truth, potentially unlimited amounts of data, and total control of scene composition, such as object pose and scene lighting. \cite{dope} demonstrates the combination of rendering on random backgrounds, as well as realistically placed objects in 3D scenes, while \cite{hinterstoisser2019annotation} shows another purely rendering-based approach. Another very common tool for data generation is BlenderProc \cite{blenderproc}, a pipeline extending Blender that allows for physically-based rendering. For a long time, the gap in feature representations between synthetic and real-world images was a problem \cite{Domain_randomization, Bridging, distributionshift}. With the use of physically-based rendering and the emergence of large foundation models such as CLIP \cite{blip2, dinov2, clip}, this problem is greatly reduced \cite{bop22}. Another problem is not to be solved that easily however: In order to render you need 3D meshes of the objects in questions. There are approaches to estimate 3D shape \cite{wonder3d} or additional views \cite{sv3d} based on a single image, but their quality is not on par with reality and their reconstruction is often purely probabilistic. Reconstructing 3D meshes with traditional methods such as photogrammetry \cite{Photogrammetry_introduction} is a very time consuming process. There is a move towards zero-shot approaches such as \cite{cnos}, which does not need images of the target objects for training, though still requires 3D meshes. 

\subsection{Significance of Reliable Data}
The use of modern DNNs in computer vision improved solutions for a large variety of challenging tasks, provided sufficient training data. A prominent modern example would be Segment Anything (SAM), which is able to segment a very large amount of different objects in the wild \cite{kirillov2023segment}. However, ML processes come with significant risks and  difficult to detect silent failures, if not handled correctly \cite{hamon2020robustness}. This is especially crucial in settings, where reliability and robustness is mandatory, such as in an industrial, clinical or dangerous contexts, which in turn can task-invalidate large unsupervised networks like SAM, due to its own limitations. One predominant challenge is, that trustworthy ML models need high-quality base data \cite{liang2022advances}. Many datasets are not task suited due to their size, inherent bias, dirtiness, or even just partial unfairness\textbackslash incorrectness \cite{whang2023data}. While these issues can be partly mitigated or worked around \cite{roh2019survey,whang2023data}, the relevant techniques introduce new layers of complexity and potential error sources. Albeit, all of these challenges can be overcome with manual labour and  diligence, this can lead to the preemptive end for startups \cite{bessen2022role} and drive up cost for large companies. 

% \subsection{Domain Gap}
% \subsubsection{Domain Randomization}
% \subsubsection{Domain Adaptation}
\subsection{Chroma Keying}
\emph{Chroma Keying} is a well established movie production technique from the 1920's, to segment objects in front of fixed mono-color backgrounds. In the beginning, a variety of different colors and shades like black/white, yellow, blue and green were popular \cite{bookgreen, bookexperimental}. Over time, the color green became the predominant background color for a variety of pragmatic reasons. It is easily distinguishable from human skin, rarely occurs outside of nature, does not require much lighting and is favourable for modern digital camera sensors because of higher sensitivity. However, some significant challenges remain: Green objects lead to falsely segmented foreground, which can also happen as a byproduct of color-bleeding or reflection. These effects strongly reduce segmentation quality for reflective materials, such as in metallic, transparent, and bright objects. Furthermore, object borders can become fuzzy due to lighting and merging effects with the background. While there are techniques to remedy drawbacks of conventional green screen, like color-unmixing \cite{aksoy2016interactive}, specialised capturing processes \cite{block2022image}, color spill neutralization \cite{grundhofer2010color}, threshold optimization \cite{phoka2017fine} and multiple background colors \cite{smith1996blue}, they open up new problems and do not yet completely solve the inherent challenges while introducing additional capturing and/or post-processing effort.

\subsection{Differentiation from similar works}
LeCun et al. \cite{lecun2004learning} use a gray turn table to record objects and rely on chroma keying to replace fore- and background, while Dirr et al. use a white background \cite{cutPaste}. 
Though a brighter background, like grey or white, would also satisfy the idea of luminance keying, they have major drawbacks in comparison to 99,99\% light absorption black. First of all, they introduce far greater challenges for correct lighting conditions, arising from background shadows, background reflections and possible fuzzy edges, due to light scattering. Furthermore, it is harder to differentiate brighter object colors with bright backgrounds. In some cases, e.g. metallic or transparent objects, a non-black background is also prone to light shining partly through an object, reflections and refractions. These challenges are all naturally solved by our proposed solution, as no light is reflected or visible on the background by any relevant measure.
Knauthe et al. use a capturing system, which foregoes a background for a white light source in a turn-table setup, which diminishes some of the challenges of a normal white background \cite{knauthe2022alignment}. However, this setup is very constrained in its rotation invariant use-case, due to the difficulties of building suitable shaped large light backgrounds. Agata et al. introduce dual color checker pattern backgrounds \cite{agata2007chroma, yamashita2008every}.
While this approach solves foreground\textbackslash background color confusion, the other issues presented earlier still persist. Additionally, the method requires more specialized backgrounds and introduces further complexity in the processing step, such as parameter tuning. Jin et al. developed a deep learning method for automatic real-time green screen keying \cite{jin2022automatic}. However, it is still limited by shadows and green spilling and requires a deep learning method with all inherent benefits and challenges. 

\section{Approach}
\subsection{Chroma and Luminance Key}
We propose a simple yet effective data recording process: we place the objects on a very low-reflectance cloth, and record around 1-minute long video clips with a smartphone. These clips are processed via a cut and paste process, that uses the easily masked objects and places them on random backgrounds. Details on the processing are discussed in section \ref{sec:dataGen}.

\subsubsection{Chroma Key with Green Screen}
\label{sec:green screen}
For comparison, we include recordings and experiments with the more common chroma keying approach. Arguably the most common colors are blue and green, with green being the most widely used color. Therefore we include a green screen in our evaluation.

\subsubsection{Luminance Key with Black Screen}
We propose Luminance Keying for fast and straightforward data acquisition for machine learning. This technique utilizes the brightness difference between an object and the background, instead of a designated color as with chroma keying. In practice, this is possible due to a textile background, which absorbs 99,99\% of visible light. This leads to high contrast between object and background, even for very dark objects, allowing for high quality masking as illustrated in Figure \ref{fig:ycbv_objects}. The recording process is exactly the same as with the conventional green screen described in section \ref{sec:green screen}.

\begin{figure*}
    \centering
    \begin{subfigure}{0.245\textwidth}
        \centering
        \includegraphics[width=\linewidth]{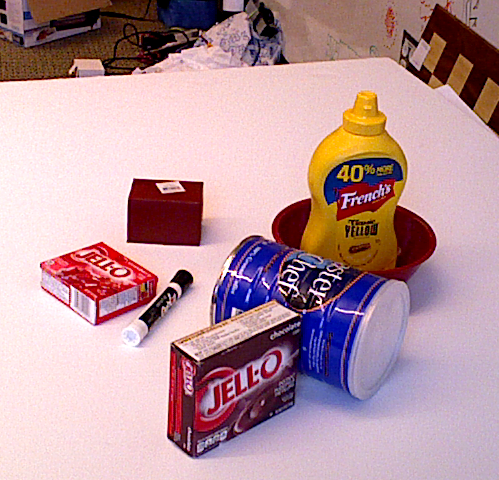}
        \caption{\textbf{REAL}}
    \end{subfigure}
    \hfill
    \begin{subfigure}{0.245\textwidth}
        \centering
        \includegraphics[width=\linewidth]{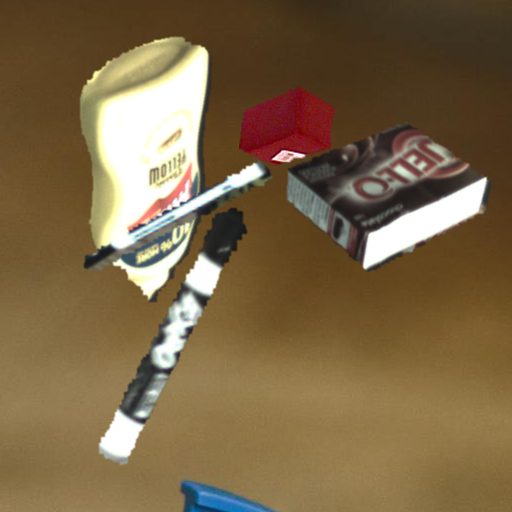}
        \caption{\textbf{RBG}}
    \end{subfigure}
    \hfill
    \begin{subfigure}{0.245\textwidth}
        \centering
        \includegraphics[width=\linewidth]{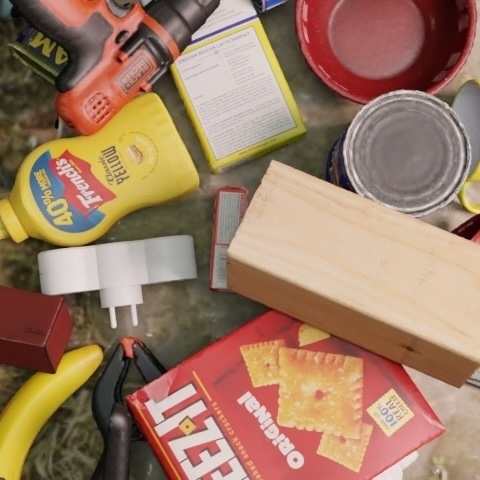}
        \caption{\textbf{PBR}}
    \end{subfigure}
    \hfill
    \begin{subfigure}{0.245\textwidth}
        \centering
        \includegraphics[width=\linewidth]{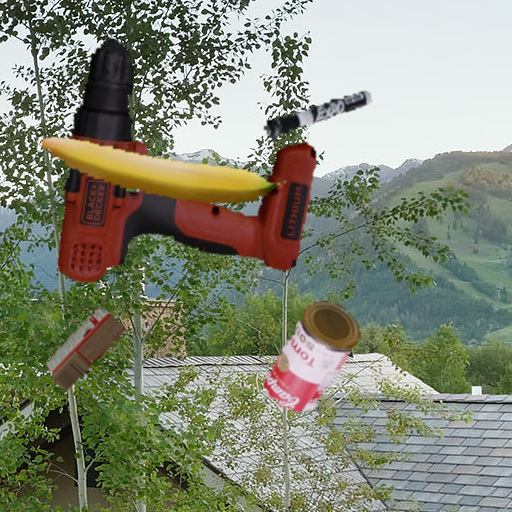}
        \caption{\textbf{PBG}}
    \end{subfigure}
    
    \vspace{0.5cm}
    
    \begin{subfigure}{0.245\textwidth}
        \centering
        \includegraphics[width=\linewidth]{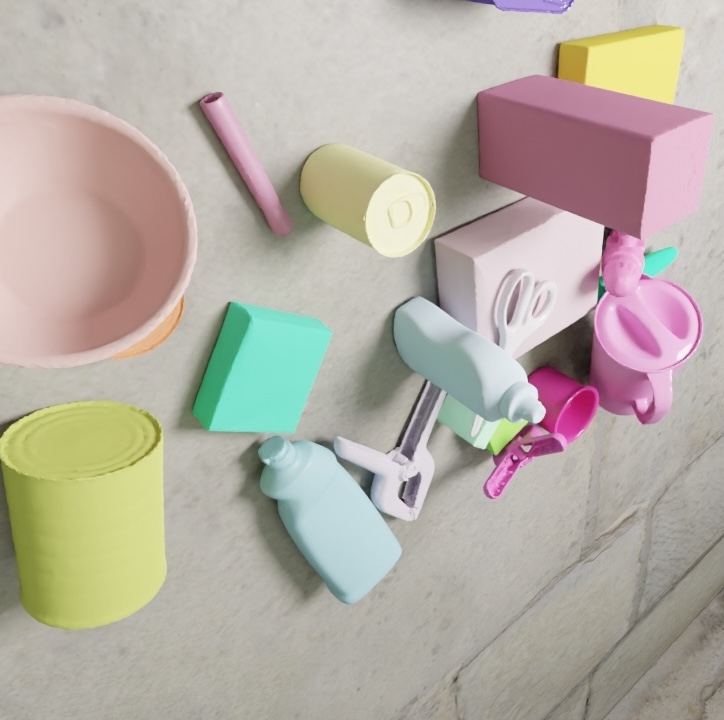}
        \caption{\textbf{PBR-rTex}}
    \end{subfigure}
    %\hfill
    \begin{subfigure}{0.245\textwidth}
        \centering
        \includegraphics[width=\linewidth]{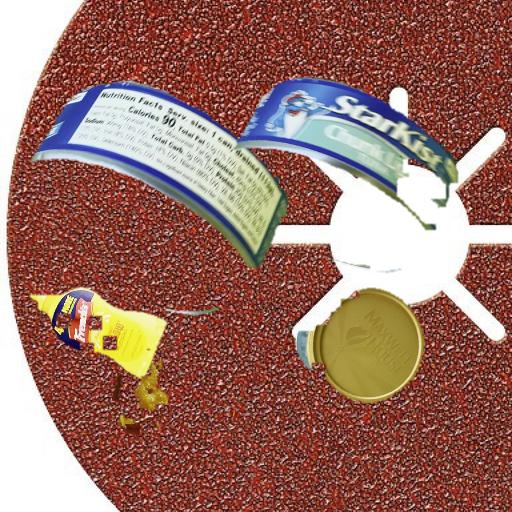}
        \caption{\textbf{CHROMA}}
    \end{subfigure}
    %\hfill
    \begin{subfigure}{0.245\textwidth}
        \centering
        \includegraphics[width=\linewidth]{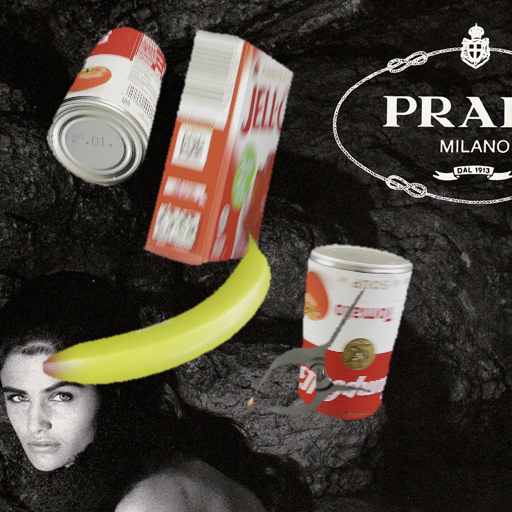}
        \caption{\textbf{LUMA} (Ours)}
    \end{subfigure}
    
    \caption{Samples from a subset of our evaluated training datasets. \textbf{REAL} are real images, \textbf{RBG} are real images with replaced backgrounds, \textbf{PBR} are physical based renderings, \textbf{PBG} are physical based renderings with background replacement, \textbf{PBR-rTex} are PBRs with randomized textures and \textbf{CHROMA}, as well as \textbf{LUMA} (Ours) stand for different capturing methods.}
    \label{fig:data}
\end{figure*}

\subsection{Baselines and Experiments}
For our evaluation we include different data sources, all visualized in Figure \ref{fig:data}: First, we include the real-world recordings of the YCB-V dataset as used in the prominent BOP Challenge \cite{bop22}. However, there is a problem with the YCB-V test images: one could argue, that the limited number of scenes and camera poses reduces the expressiveness in terms of generalization. Therefore we also evaluate on the physically-based rendering (pbr) dataset provided by BOP. This dataset uses the reconstructed textures, leading to a high similarity in appearance compared to the real images, but has a much wider range of camera poses, object configurations, and lighting situations. These two datasets, namely REAL and PBR, are used as a baseline on which all other data sources have to be tested. Also, we test all of our approaches on in-distribution samples, that is, on disjoint splits from the dataset, i.e. when training on our luminance images, we test on REAL, PBR, as well as on a disjoint set of luminance images. This should serve as a measure on how well an experiment generalizes to out-of-distribution samples. \\
In more detail we use the following data representations:\\
\textbf{Real-world images (REAL).} Real-world recordings as described above. \\
\textbf{Real-world images with background replacement (RBG).} Real-world recordings. To test the influence of our background replacement script, we apply it to the real images as well. This gives us an idea on how much performance we loose because of our simple crop and paste approach of data generation. \\
\textbf{Physically-based renderings (PBR).} PBR renderings with reconstructed textures, realistic object placement, and light transport, as well as a high degree of camera and object pose variation.  \\
\textbf{Physically-based renderings with background replacement (PBG).} Same as above, but again using background replacement for measuring its influence on performance. \\
\textbf{Physically-based renderings with randomized textures (PBR-rTex).} To simulate cases in which geometry is available, though we lack realistic textures and materials, we include a set of pbr images with randomized surface attributes. Namely we randomize texture and surface details such as reflectance behavior. \\
\textbf{Chroma key images using green screen (CHROMA).} A set of green screen recordings to illustrate what results one can expect from using run-of-the-mill chroma keying methods. We use the green screen for segmentation and paste the crops on random photographs. \\
\textbf{Luminance key images using black screen (LUMA).} Our proposed method. A set of recordings utilizing the low-reflection black screen. Again, we have to crop and paste the objects on random backgrounds using our replacement script. \\

\subsection{Training}
We demonstrate the applicability of our approach on the 2D object detection use case and train YOLOX \cite{yolox} networks. \newline
\textbf{Data Generation} using cut and paste background replacement. \label{sec:dataGen}
%Explain our BG replacement script. \newline
To achieve technically good replacement results, we use a variety of probabilistic mechanisms that are applied during the replacement process inspired by \cite{cutPasteLearn}. First, based on the masks provided, we filter out all objects in an image that are overlapped by others. This is trivial for luma, because there is no overlap during the acquisition, giving us near-perfect masks. In addition, we also make sure that the objects are not cut off at the edge of the image and finally do a crop, centered around the object. Next we apply random affine transformations such as scaling, rotation and translation to the objects where the variable scaling, rotation and translation ensures variety in the appearance of the objects. After this process, the objects are placed on the background image based on randomly generated positions. We allow an overlap among objects of up to 20\% in order to come closer to the real scenarios of YCB-V and make the network learn to deal with occlusion. For bleding and to remove jaggies, the object masks are eroded and Gaussian noise is applied. As for the backgrounds we used random crops of images from the 50k HQ-data set \cite{hqImageDataset}. Random cropping prevents the same background from being used more than once. 
\textbf{Parameterization.}
To evaluate the practicability of our proposed method, we do an extensive comparison of different training sets. As shown in \cite{hinterstoisser_freezing} freezing the backbone layers is useful for reducing the impact of domain shift. Since we introduce additional domain discrepancy with our background replacement, we add freezing to our evaluation of training from scratch, and using COCO-pretrained weights for initialization. As for the parameterization of YOLOX, we stick closely to the values and process as presented in their paper \cite{yolox}. Most importantly for our comprehensive quantitative analysis we train model size "tiny" with image size 512x512, batch sizes of 64, half precision, multiscale range of 13, and for 300 epochs. %For the qualitative results we train model configuration "X" with batch size 16 and image size 1024x1024. 

\section{Results}
We report quantitative results on the common YCB-V object set and show samples of the achievable visual quality of the approach. In addition we show some problems of chroma keying, which luminance keying does not have in section \ref{sec:qualitative}. 

\subsection{Quantitative Results}

\begin{table*}
\centering
\resizebox{\textwidth}{!}{
\begin{tblr}{
  column{2} = {c},
  cell{1}{1} = {c},
  cell{1}{3} = {c},
  cell{1}{4} = {c},
  cell{1}{5} = {c},
  cell{1}{6} = {c},
  cell{1}{7} = {c},
  cell{2}{3} = {c},
  cell{2}{4} = {c},
  cell{2}{5} = {c},
  cell{2}{6} = {c},
  cell{2}{7} = {c},
  cell{3}{3} = {c},
  cell{3}{4} = {c},
  cell{3}{5} = {c},
  cell{3}{6} = {c},
  cell{3}{7} = {c},
  cell{4}{3} = {c},
  cell{4}{4} = {c},
  cell{4}{5} = {c},
  cell{4}{6} = {c},
  cell{4}{7} = {c},
  cell{5}{3} = {c},
  cell{5}{4} = {c},
  cell{5}{5} = {c},
  cell{5}{6} = {c},
  cell{5}{7} = {c},
  cell{6}{3} = {c},
  cell{6}{4} = {c},
  cell{6}{5} = {c},
  cell{6}{6} = {c},
  cell{6}{7} = {c},
  cell{7}{3} = {c},
  cell{7}{4} = {c},
  cell{7}{5} = {c},
  cell{7}{6} = {c},
  cell{7}{7} = {c},
  cell{8}{3} = {c},
  cell{8}{4} = {c},
  cell{8}{5} = {c},
  cell{8}{6} = {c},
  cell{8}{7} = {c},
  cell{9}{3} = {c},
  cell{9}{4} = {c},
  cell{9}{5} = {c},
  cell{9}{6} = {c},
  cell{9}{7} = {c},
  cell{10}{3} = {c},
  cell{10}{4} = {c},
  cell{10}{5} = {c},
  cell{10}{6} = {c},
  cell{10}{7} = {c},
  cell{11}{3} = {c},
  cell{11}{4} = {c},
  cell{11}{5} = {c},
  cell{11}{6} = {c},
  cell{11}{7} = {c},
  cell{12}{3} = {c},
  cell{12}{4} = {c},
  cell{12}{5} = {c},
  cell{12}{6} = {c},
  cell{12}{7} = {c},
  cell{13}{3} = {c},
  cell{13}{4} = {c},
  cell{13}{5} = {c},
  cell{13}{6} = {c},
  cell{13}{7} = {c},
  cell{14}{3} = {c},
  cell{14}{4} = {c},
  cell{14}{5} = {c},
  cell{14}{6} = {c},
  cell{14}{7} = {c},
  cell{15}{3} = {c},
  cell{15}{4} = {c},
  cell{15}{5} = {c},
  cell{15}{6} = {c},
  cell{15}{7} = {c},
  cell{16}{3} = {c},
  cell{16}{4} = {c},
  cell{16}{5} = {c},
  cell{16}{6} = {c},
  cell{16}{7} = {c},
  cell{17}{3} = {c},
  cell{17}{4} = {c},
  cell{17}{5} = {c},
  cell{17}{6} = {c},
  cell{17}{7} = {c},
  cell{18}{3} = {c},
  cell{18}{4} = {c},
  cell{18}{5} = {c},
  cell{18}{6} = {c},
  cell{18}{7} = {c},
  cell{19}{3} = {c},
  cell{19}{4} = {c},
  cell{19}{5} = {c},
  cell{19}{6} = {c},
  cell{19}{7} = {c},
  cell{20}{3} = {c},
  cell{20}{4} = {c},
  cell{20}{5} = {c},
  cell{20}{6} = {c},
  cell{20}{7} = {c},
  cell{21}{3} = {c},
  cell{21}{4} = {c},
  cell{21}{5} = {c},
  cell{21}{6} = {c},
  cell{21}{7} = {c},
  cell{22}{3} = {c},
  cell{22}{4} = {c},
  cell{22}{5} = {c},
  cell{22}{6} = {c},
  cell{22}{7} = {c},
  cell{23}{3} = {c},
  cell{23}{4} = {c},
  cell{23}{5} = {c},
  cell{23}{6} = {c},
  cell{23}{7} = {c},
  cell{2}{8} = {c},
  cell{3}{8} = {c},
  cell{4}{8} = {c},
  cell{5}{8} = {c},
  cell{6}{8} = {c},
  cell{7}{8} = {c},
  cell{8}{8} = {c},
  cell{9}{8} = {c},
  cell{10}{8} = {c},
  cell{11}{8} = {c},
  cell{12}{8} = {c},
  cell{13}{8} = {c},
  cell{14}{8} = {c},
  cell{15}{8} = {c},
  cell{16}{8} = {c},
  cell{17}{8} = {c},
  cell{18}{8} = {c},
  cell{19}{8} = {c},
  cell{20}{8} = {c},
  cell{21}{8} = {c},
  cell{22}{8} = {c},
  cell{23}{8} = {c},
  cell{24}{8} = {c},
  vline{2} = {-}{},
  hline{1-2,23,25} = {-}{},
}\textbf{}
TRAIN SET:               & PBR-rTex & PBR               & PBG            & REAL              & RBG           & CHROMA           & LUMA (Ours)       \\
002\_master\_chef\_can*  & 0.19/\textbf{1.47}/-         & 44.65/\textbf{62.13}/44.65 & 21.74/\textbf{53.56}/17.35 & 6.29/\textbf{73.76}/73.76  & 2.2/\textbf{2.84}/0.0      & 9.54/\textbf{46.75}/1.56  & 11.74/\textbf{46.17/}3.64 \\
003\_cracker\_box~       & 0.09/\textbf{0.0}/-          & 46.86/\textbf{20.65}/46.86 & 16.33/\textbf{5.15}/14.3   & 12.54/\textbf{43.34}/43.34 & 10.06/\textbf{9.65}/0.0    & 5.31/\textbf{10.2}/0.02   & 13.68/\textbf{20.9}/0.51  \\
004\_sugar\_box~         & 0.23/\textbf{0.02}/-         & 32.95/\textbf{48.36}/32.95 & 11.77/\textbf{44.75}/13.03 & 5.62/\textbf{49.54}/49.54  & 9.37/\textbf{9.26}/66.26   & 9.42/\textbf{38.23}/0.13  & 14.97/\textbf{52.64}/0.48 \\
005\_tomato\_soup\_can   & 0.08/\textbf{0.3}/-          & 38.04/\textbf{55.53}/38.04 & 9.59/\textbf{59.13}/10.47  & 5.75/\textbf{70.21}/70.21  & 9.44/\textbf{9.43}/78.67   & 4.73/\textbf{36.42}/0.46  & 10.27/\textbf{29.17}/1.42 \\
006\_mustard\_bottle     & 0.65/\textbf{0.14}/-         & 38.86/\textbf{78.2}/38.86  & 14.19/\textbf{76.07}/7.38  & 10.68/\textbf{88.82}/88.82 & 12.7/\textbf{12.45}/9.72   & 16.75/\textbf{73.42}/2.48 & 17.63/\textbf{79.9}/5.3   \\
007\_tuna\_fish\_can     & 0.24/\textbf{0.2}/-          & 45.98/\textbf{70.77}/45.98 & 11.04/\textbf{65.36}/19.02 & 12.55/\textbf{79.35}/79.35 & 10.69/\textbf{10.74}/72.85 & 5.8/\textbf{53.93}/0.23   & 11.34/\textbf{60.97}/1.15 \\\textbf{}
008\_pudding\_box~       & 0.11/\textbf{0.0}/-          & 28.54/\textbf{4.48}/28.54  & 17.19/\textbf{1.2}/22.23   & 1.77/\textbf{6.64}/6.64    & 9.88/\textbf{9.84}/0.0     & 4.05/\textbf{2.43}/1.16   & 5.46/\textbf{0.92}/2.5    \\
009\_gelatin\_box~       & 0.11/\textbf{0.0}/-          & 36.82/\textbf{71.21}/36.82 & 15.61/\textbf{5.73}/22.94  & 6.11/\textbf{65.41}/65.41  & 11.61/\textbf{11.12}/66.94 & 2.26/\textbf{0.52}/0.6    & 1.33/\textbf{0.55}/2.41   \\
010\_potted\_meat\_can   & 0.31/\textbf{0.27}/-         & 36.28/\textbf{37.64}/36.28 & 11.4/\textbf{18.3}/11.33   & 4.04/\textbf{54.23}/54.23  & 9.06/\textbf{9.04}/40.73   & 1.29/\textbf{0.27}/2.83   & 11.52/\textbf{26.22}/0.88 \\
011\_banana~             & 5.79/\textbf{6.58}/-         & 34.84/\textbf{54.24}/34.84 & 14.91/\textbf{40.67}/20.34 & 15.62/\textbf{50.18}/50.18 & 14.18/\textbf{14.28}/88.81 & 17.17\textbf{/51.06}/3.16 & 14.95/\textbf{48.64}/0.63 \\
019\_pitcher\_base*      & 4.27/\textbf{3.12}/-         & 54.15/\textbf{55.57}/54.15 & 1.08/\textbf{29.29}/0.55   & 13.56/\textbf{80.89}/80.89 & 0.0/\textbf{0.0}/0.0       & 6.68/\textbf{0.38}/0.67   & 6.28/\textbf{0.32}/0.09   \\
021\_bleach\_cleanser    & 0.92/\textbf{2.56}/-         & 33.82/\textbf{43.58}/33.82 & 8.23/\textbf{36.19}/6.22   & 2.42/\textbf{61.31}/61.31  & 2.09/\textbf{2.15}/8.35    & 8.03/\textbf{47.44}/1.56  & 10.53/\textbf{54.61}/5.26 \\
024\_bowl~               & 2.14/\textbf{0.22}/-         & 54.84/\textbf{46.75}/54.84 & 4.96/\textbf{0.01}/5.24    & 9.04/\textbf{56.9}/56.9    & 2.47/\textbf{1.93}/0.0     & 0.06/\textbf{0.21}/0.08   & 6.99/\textbf{14.44}/1.3   \\
025\_mug~                & 0.18/\textbf{3.3}/-          & 46.37/\textbf{58.14}/46.37 & 10.89/\textbf{1.87}/1.72   & 4.0/\textbf{67.35}/67.35   & 3.01/\textbf{3.01}/0.0     & 5.45/\textbf{31.17}/1.02  & 2.04/\textbf{14.37}/0.05  \\
035\_power\_drill*~      & 0.3/\textbf{0.0}/-           & 23.28/\textbf{13.26}/23.28 & 3.24/\textbf{0.76}/1.39    & 2.28/\textbf{43.72}/43.72  & 0.0/\textbf{0.0}/0.0       & 3.87/\textbf{5.71}/0.03   & 4.46/\textbf{13.23}/0.04  \\
036\_wood\_block         & 2.1/\textbf{0.37}/-          & 45.99/\textbf{24.57}/45.99 & 15.57/\textbf{8.22}/14.56  & 8.42/\textbf{34.06}/34.06  & 5.85/\textbf{5.8}/0.0      & 10.28/\textbf{12.75}/0.64 & 14.1/\textbf{17.32}/2.53  \\
037\_scissors~           & 0.67/\textbf{0.0}/-          & 19.32/\textbf{5.48}/19.32  & 0.67/\textbf{0.16}/3.42    & 5.98/\textbf{1.96}/1.96    & 1.36/\textbf{1.32}/0.0     & 1.4/\textbf{0.33}/0.69    & 1.35/\textbf{0.52}/0.01   \\
040\_large\_marker       & 0.04/\textbf{0.0}/-          & 16.1/\textbf{47.65}/16.1   & 2.54/\textbf{22.2}/9.52    & 1.83/\textbf{56.73}/56.73  & 0.47/\textbf{0.41}/8.58    & 0.87/\textbf{31.87}/0.23  & 0.16/\textbf{5.42}/0.4    \\
051\_large\_clamp        & 0.85/\textbf{0.0}/-          & 18.2/\textbf{47.43}/18.2   & 0.07/\textbf{0.25}/2.5     & 2.96/\textbf{6.4}/6.4      & 0.07/\textbf{0.09}/0.01    & 1.85/\textbf{0.28}/2.01   & 2.93/\textbf{17.75}/0.07  \\
052\_extra\_large\_clamp & 0.38/\textbf{0.17}/-         & 19.27/\textbf{7.21}/19.27  & 0.01/\textbf{0.0}/2.45     & 2.38/\textbf{9.8}/9.8      & 0.5/\textbf{0.08}/0.0      & 8.79/\textbf{11.31}/1.13  & 2.95/\textbf{0.29}/1.82   \\
061\_foam\_brick~        & 0.71/\textbf{0.01}/-         & 36.94/\textbf{55.61}/36.94 & 15.86/\textbf{3.38}/24.54  & 8.09/\textbf{67.82}/67.82  & 6.99/\textbf{6.63}/0.0     & 6.43/\textbf{1.64}/1.11   & 6.01/\textbf{5.22}/0.91   \\
AVERAGE YCB-V            & 0.97/\textbf{0.89}/-         & 35.81/\textbf{43.26}/35.81 & 9.85/\textbf{22.49}/10.98  & 6.76/\textbf{50.88}/50.88  & 5.81/\textbf{5.72}/21.0    & 6.19/\textbf{21.73}/1.04  & 8.13/\textbf{24.27}/1.49  \\
AVERAGE YCB-V18          & 0.87/\textbf{0.78}/-         & 35.0/\textbf{43.19}/35.0   & 10.05/\textbf{21.59}/11.73 & 6.66/\textbf{48.34}/48.34  & 6.65/\textbf{6.51}/24.5    & 6.11/\textbf{22.42}/1.09  & 8.23/\textbf{24.99}/1.54  
\end{tblr}
}
\caption{AP (higher is better) of all evaluated data representations, each tested on PBR, REAL, and in-distribution. In-distribution means the test set stems from the same data source (e.g. photograph, rendering, crop and paste) as the training set. Results on REAL in \textbf{bold font}. As expected, we see best results when training on the train split of the real data (REAL), followed by the physically simulated 3D scenes (PBR). Most importantly, among all 4 methods using the cut and paste approach for background replacement (PBG, RBG, CHROMA, LUMA), luminance (our proposed method) outperforms all others when tested on the real test data (REAL).}
\label{tab:ap}
\end{table*}

\begin{table*}
\centering
\resizebox{\textwidth}{!}{
\begin{tblr}{
  column{2} = {c},
  cell{1}{1} = {c},
  cell{1}{3} = {c},
  cell{1}{4} = {c},
  cell{1}{5} = {c},
  cell{1}{6} = {c},
  cell{1}{7} = {c},
  cell{2}{3} = {c},
  cell{2}{4} = {c},
  cell{2}{5} = {c},
  cell{2}{6} = {c},
  cell{2}{7} = {c},
  cell{3}{3} = {c},
  cell{3}{4} = {c},
  cell{3}{5} = {c},
  cell{3}{6} = {c},
  cell{3}{7} = {c},
  cell{4}{3} = {c},
  cell{4}{4} = {c},
  cell{4}{5} = {c},
  cell{4}{6} = {c},
  cell{4}{7} = {c},
  cell{5}{3} = {c},
  cell{5}{4} = {c},
  cell{5}{5} = {c},
  cell{5}{6} = {c},
  cell{5}{7} = {c},
  cell{6}{3} = {c},
  cell{6}{4} = {c},
  cell{6}{5} = {c},
  cell{6}{6} = {c},
  cell{6}{7} = {c},
  cell{7}{3} = {c},
  cell{7}{4} = {c},
  cell{7}{5} = {c},
  cell{7}{6} = {c},
  cell{7}{7} = {c},
  cell{8}{3} = {c},
  cell{8}{4} = {c},
  cell{8}{5} = {c},
  cell{8}{6} = {c},
  cell{8}{7} = {c},
  cell{9}{3} = {c},
  cell{9}{4} = {c},
  cell{9}{5} = {c},
  cell{9}{6} = {c},
  cell{9}{7} = {c},
  cell{10}{3} = {c},
  cell{10}{4} = {c},
  cell{10}{5} = {c},
  cell{10}{6} = {c},
  cell{10}{7} = {c},
  cell{11}{3} = {c},
  cell{11}{4} = {c},
  cell{11}{5} = {c},
  cell{11}{6} = {c},
  cell{11}{7} = {c},
  cell{12}{3} = {c},
  cell{12}{4} = {c},
  cell{12}{5} = {c},
  cell{12}{6} = {c},
  cell{12}{7} = {c},
  cell{13}{3} = {c},
  cell{13}{4} = {c},
  cell{13}{5} = {c},
  cell{13}{6} = {c},
  cell{13}{7} = {c},
  cell{14}{3} = {c},
  cell{14}{4} = {c},
  cell{14}{5} = {c},
  cell{14}{6} = {c},
  cell{14}{7} = {c},
  cell{15}{3} = {c},
  cell{15}{4} = {c},
  cell{15}{5} = {c},
  cell{15}{6} = {c},
  cell{15}{7} = {c},
  cell{16}{3} = {c},
  cell{16}{4} = {c},
  cell{16}{5} = {c},
  cell{16}{6} = {c},
  cell{16}{7} = {c},
  cell{17}{3} = {c},
  cell{17}{4} = {c},
  cell{17}{5} = {c},
  cell{17}{6} = {c},
  cell{17}{7} = {c},
  cell{18}{3} = {c},
  cell{18}{4} = {c},
  cell{18}{5} = {c},
  cell{18}{6} = {c},
  cell{18}{7} = {c},
  cell{19}{3} = {c},
  cell{19}{4} = {c},
  cell{19}{5} = {c},
  cell{19}{6} = {c},
  cell{19}{7} = {c},
  cell{20}{3} = {c},
  cell{20}{4} = {c},
  cell{20}{5} = {c},
  cell{20}{6} = {c},
  cell{20}{7} = {c},
  cell{21}{3} = {c},
  cell{21}{4} = {c},
  cell{21}{5} = {c},
  cell{21}{6} = {c},
  cell{21}{7} = {c},
  cell{22}{3} = {c},
  cell{22}{4} = {c},
  cell{22}{5} = {c},
  cell{22}{6} = {c},
  cell{22}{7} = {c},
  cell{23}{3} = {c},
  cell{23}{4} = {c},
  cell{23}{5} = {c},
  cell{23}{6} = {c},
  cell{23}{7} = {c},
  vline{2} = {-}{},
  cell{2}{8} = {c},
  cell{3}{8} = {c},
  cell{4}{8} = {c},
  cell{5}{8} = {c},
  cell{6}{8} = {c},
  cell{7}{8} = {c},
  cell{8}{8} = {c},
  cell{9}{8} = {c},
  cell{10}{8} = {c},
  cell{11}{8} = {c},
  cell{12}{8} = {c},
  cell{13}{8} = {c},
  cell{14}{8} = {c},
  cell{15}{8} = {c},
  cell{16}{8} = {c},
  cell{17}{8} = {c},
  cell{18}{8} = {c},
  cell{19}{8} = {c},
  cell{20}{8} = {c},
  cell{21}{8} = {c},
  cell{22}{8} = {c},
  cell{23}{8} = {c},
  cell{24}{8} = {c},
  hline{1-2,23,25} = {-}{},
}
TRAIN SET:               & PBR-rTex & PBR               & PBG            & REAL              & RBG           & CHROMA            & LUMA (Ours)        \\
002\_master\_chef\_can*  & 8.05/\textbf{18.8}/-         & 63.67/\textbf{77.07}/63.67 & 39.66/\textbf{71.67}/54.88 & 14.9/\textbf{79.0}/79.0    & 4.59/\textbf{4.63}/0.0     & 34.94/\textbf{74.8}/9.17   & 33.24/\textbf{71.97}/16.74 \\
003\_cracker\_box~       & 7.51/\textbf{0.22}/-         & 62.82/\textbf{66.44}/62.82 & 34.83/\textbf{49.02}/56.82 & 22.93/\textbf{62.89}/62.89 & 15.5/\textbf{21.64}/0.0    & 13.62/\textbf{36.22}/1.14  & 26.64/\textbf{48.53}/4.63  \\
004\_sugar\_box~         & 6.75/\textbf{2.61}/-         & 53.39/\textbf{66.8}/53.39  & 31.07/\textbf{74.37}/48.63 & 18.53/\textbf{61.63}/61.63 & 27.76/\textbf{75.68}/77.14 & 22.13/\textbf{52.69}/5.68  & 32.84/\textbf{69.07}/7.23  \\
005\_tomato\_soup\_can   & 4.2/\textbf{5.96}/-          & 54.06/\textbf{69.71}/54.06 & 22.89/\textbf{66.91}/35.95 & 20.81/\textbf{75.09}/75.09 & 26.39/\textbf{68.96}/84.03 & 25.11/\textbf{54.96}/7.86  & 30.4/\textbf{61.8}/14.32   \\
006\_mustard\_bottle     & 11.46/\textbf{8.27}/-        & 57.22/\textbf{87.6}/57.22  & 27.01/\textbf{83.87}/34.08 & 23.68/\textbf{91.27}/91.27 & 28.38/\textbf{82.33}/55.56 & 30.54/\textbf{83.87}/13.41 & 32.17/\textbf{84.13}/21.22 \\
007\_tuna\_fish\_can     & 7.29/\textbf{4.37}/-         & 57.74/\textbf{78.83}/57.74 & 26.17/\textbf{72.1}/51.98  & 30.23/\textbf{83.77}/83.77 & 30.72/\textbf{75.17}/84.8  & 24.34/\textbf{72.77}/7.5   & 24.88/\textbf{73.3}/11.4   \\
008\_pudding\_box~       & 3.31/\textbf{0.27}/-         & 48.6/\textbf{56.4}/48.6    & 33.25/\textbf{23.73}/65.56 & 14.17/\textbf{48.93}/48.93 & 24.87/\textbf{54.27}/0.0   & 26.21/\textbf{43.6}/8.24   & 27.76/\textbf{27.07}/10.97 \\
009\_gelatin\_box~       & 2.96/\textbf{0.0}/-          & 49.85/\textbf{78.4}/49.85  & 26.41/\textbf{50.8}/60.06  & 21.31/\textbf{74.27}/74.27 & 28.77/\textbf{72.4}/85.1   & 10.21/\textbf{23.87}/4.89  & 8.33/\textbf{16.27}/9.05   \\
010\_potted\_meat\_can   & 6.69/\textbf{0.84}/-         & 53.21/\textbf{64.98}/53.21 & 28.8/\textbf{48.09}/38.76  & 19.73/\textbf{59.6}/59.6   & 28.64/\textbf{54.31}/78.69 & 7.47/\textbf{10.4}/7.67    & 26.99/\textbf{52.93}/5.43  \\
011\_banana~             & 26.99/\textbf{31.8}/-        & 49.23/\textbf{70.87}/49.23 & 24.18/\textbf{55.27}/46.25 & 28.16/\textbf{61.4}/61.4   & 23.91/\textbf{54.07}/90.0  & 25.33/\textbf{58.27}/15.58 & 22.91/\textbf{55.33}/10.21 \\
019\_pitcher\_base*      & 29.3/\textbf{28.53}/-        & 71.89/\textbf{77.33}/71.89 & 4.47/\textbf{34.76}/9.04   & 26.38/\textbf{84.22}/84.22 & 0.0/\textbf{0.0}/0.0       & 27.79/\textbf{8.18}/5.28   & 25.67/\textbf{6.44}/3.86   \\
021\_bleach\_cleanser    & 16.59/\textbf{17.13}/-       & 54.95/\textbf{60.43}/54.95 & 22.5/\textbf{58.4}/33.01   & 13.72/\textbf{71.0}/71.0   & 2.78/\textbf{17.27}/24.38  & 20.38/\textbf{59.2}/11.14  & 20.94/\textbf{63.1}/13.81  \\
024\_bowl~               & 25.35/\textbf{9.47}/-        & 66.4/\textbf{73.07}/66.4   & 16.25/\textbf{1.13}/29.93  & 14.96/\textbf{63.13}/63.13 & 2.67/\textbf{18.47}/0.0    & 0.58/\textbf{5.0}/4.0      & 12.03/\textbf{55.6}/\textbf{}11.33  \\
025\_mug~                & 7.74/\textbf{29.13}/-        & 60.39/\textbf{74.13}/60.39 & 17.04/\textbf{11.47}/22.32 & 18.81/\textbf{71.93}/71.93 & 15.77/\textbf{72.2}/0.0    & 19.77/\textbf{72.53}/10.26 & 6.37/\textbf{45.8}/2.7     \\
035\_power\_drill*~      & 7.57/\textbf{0.07}/-         & 47.82/\textbf{44.43}/47.82 & 11.34/\textbf{7.3}/25.11   & 14.38/\textbf{54.23}/54.23 & 0.0/\textbf{0.0}/0.0       & 24.05/\textbf{41.73}/1.92  & 20.71/\textbf{54.23}/0.96  \\
036\_wood\_block         & 15.6/\textbf{19.87}/-        & 62.63/\textbf{65.73/}62.63 & 32.68/\textbf{32.93}/46.97 & 20.44/\textbf{50.8}/50.8   & 12.61/\textbf{28.67}/0.0   & 26.27/\textbf{56.67}/8.75  & 28.67/\textbf{57.07}/9.82  \\
037\_scissors~           & 7.87\textbf{/0.27}/-         & 34.22/\textbf{20.8}/34.22  & 4.53/\textbf{1.73}/35.71   & 12.38/\textbf{6.4}/6.4     & 9.77/\textbf{8.27}/0.0     & 9.97/\textbf{6.27}/3.09    & 7.06/\textbf{3.73}/1.22    \\
040\_large\_marker       & 2.72/\textbf{0.13}/-         & 30.6/\textbf{62.0}/30.6    & 8.85/\textbf{39.67}/38.86  & 7.12/\textbf{63.33}/63.33  & 8.13/\textbf{60.53}/33.79  & 6.54/\textbf{59.8}/5.68    & 3.54/\textbf{31.47}/3.1    \\
051\_large\_clamp        & 6.62/\textbf{0.0}/-          & 43.2/\textbf{69.8}/43.2    & 3.06/\textbf{12.47}/28.43  & 10.62/\textbf{32.8}/32.8   & 0.35/\textbf{11.47}/3.33   & 12.01/\textbf{9.8}/10.44   & 13.17/\textbf{36.53}/2.86  \\
052\_extra\_large\_clamp & 8.39/\textbf{4.2}/-          & 47.61/\textbf{45.87}/47.61 & 0.77/\textbf{0.0}/30.26    & 9.41/\textbf{35.87}/35.87  & 0.07/\textbf{0.07}/0.0     & 21.33/\textbf{26.93}/6.18  & 12.54/\textbf{9.07}/4.32   \\
061\_foam\_brick~        & 6.92/\textbf{3.47}/-         & 52.27/\textbf{72.93}/52.27 & 25.99/\textbf{59.87}/56.78 & 23.88/\textbf{76.27}/76.27 & 19.89/\textbf{71.73}/0.0   & 22.89/\textbf{25.47}/4.77  & 22.8/\textbf{54.4}/6.92    \\
AVERAGE YCB-V            & 10.47\textbf{}/\textbf{8.83}/-        & 53.42/\textbf{65.89}/53.42 & 21.04/\textbf{40.74}/40.45 & 18.41/\textbf{62.28}/62.28 & 14.84/\textbf{40.58}/29.37 & 19.6/\textbf{42.05}/7.27   & 20.93/\textbf{46.56}/\textbf{}8.2   \\
AVERAGE YCB-V18          & 9.72/\textbf{7.67}/-         & 52.13/\textbf{65.82}/52.13 & 21.46/\textbf{41.21}/42.24 & 18.38/\textbf{60.58}/60.58 & 17.05/\textbf{47.08}/34.27 & 18.04/\textbf{42.13}/7.57  & 20.0/\textbf{46.96}/8.36   
\end{tblr}
}
\caption{AR (higher is better) of all evaluated data representation, corresponding to Table \ref{tab:ap}. Results on REAL in \textbf{bold font}. LUMA (Ours) compares favourably, outperforming PBG, RBG, and CHROMA.}
\label{tab:ar}
\end{table*}

In our first experiments we looked into answering the question, whether we want to train from scratch or load pre-trained weights, as well as whether to freeze the backbone. For a comprehensive list of all results, especially for unfreezed backbones, please refer to our tables presented in the appendix. Here we report results when freezing the backbone, as these led to best performance with our cut and paste background replacement. For evaluation we report both the Average Precision (AP) and the Average Recall (AR) metric in Tables \ref{tab:ap} and \ref{tab:ar}. Since the currently available set of YCB-V objects has three modified objects (002\_master\_chef\_can has a modified texture, for 019\_pitcher\_base the color changed from blue to a transparent color with a red lid, and for 035\_power\_drill there are some slight changes to the model), we report results on the subset of unchanged objects and call it YCB-V18, as well as the results of the complete set.\\
As expected training and evaluating on REAL lead to best performance overall. This is no surprise since, among all data representations tested, features in the REAL training set are closest to the features found in the REAL test set. Physically-based rendering (PBR) comes in second place. The good performance of PBR can be explained with the photo-realistic depiction, wide range of physically-correct object placements, and lighting variations. Surprisingly the PBR set with randomized textures has very poor performance, far worse than expected. While we assumed a big drop, the extent might have to do with model size and might be less pronounced with more trainable parameters. As the 4 remaining approaches all build upon the cut and paste approach, we can directly compare the quality of the segmentation, as well as image fidelity. Here we find an outlier in the PBG (real photographs with background replacement) set: it under-performs the other approaches by a wide margin. Our assumption is that the reason is the comparatively poor masking quality. Masks are not pixel perfect and sometimes include background pixels or pixels belonging to an occluding object. Mask quality is no problem for RBG (pbr images with background replacement), having perfect masks by virtue of complete scene knowledge, but we argue RBG is held back by other problems, some similar to the ones found with chroma keying with color bleeding being the most obvious. Realistic light transport is a feature in fully rendered scenes, but becomes a detriment with cutting and pasting. Also, the textures, while being high fidelity, are not as good as real photographs. Finally, comparing LUMA with CHROMA we see an outperformance of more than 11\% on both YCB-V18, as well as on the complete YCB-V set. 

\begin{figure}
    \centering
    \begin{minipage}{0.235\textwidth}
        \centering
        \includegraphics[width=\linewidth]{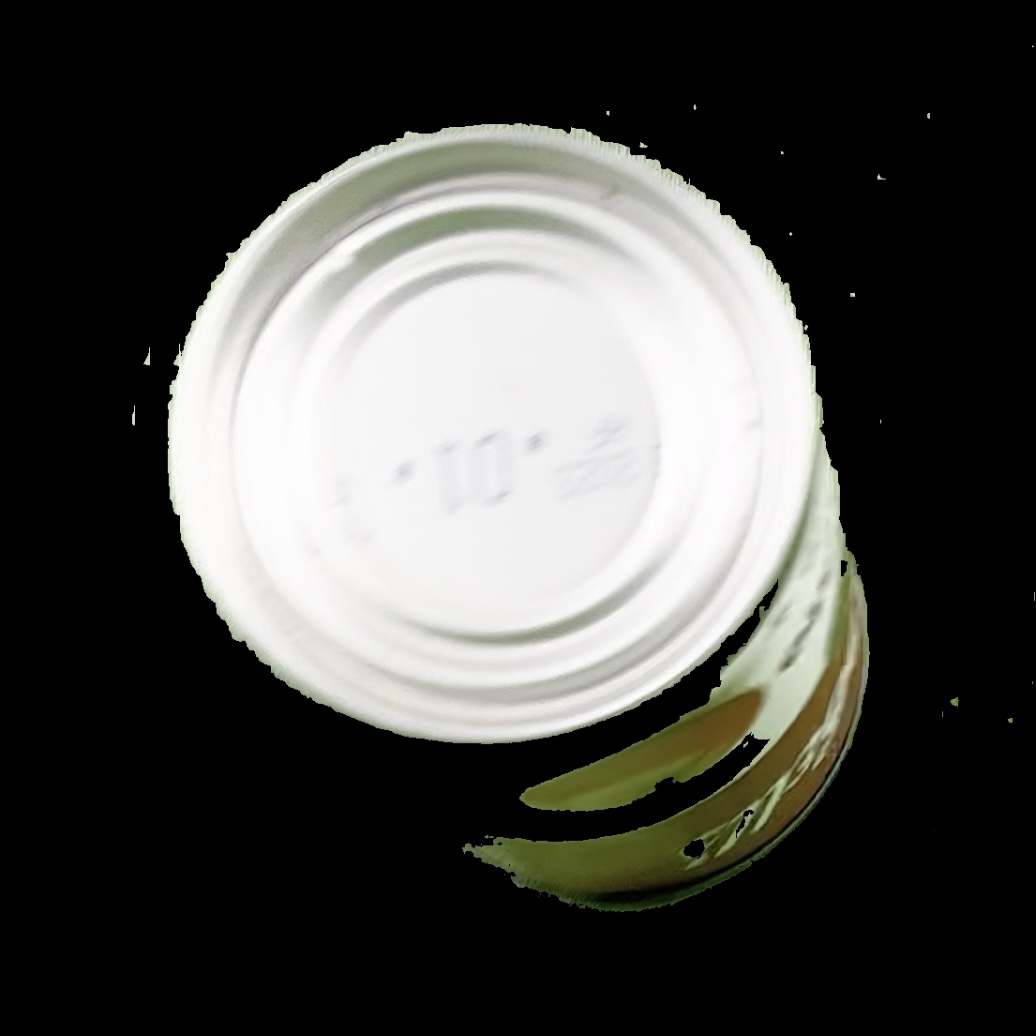}
    \end{minipage}
    \hfill
    \begin{minipage}{0.235\textwidth}
        \centering
        \includegraphics[width=\linewidth]{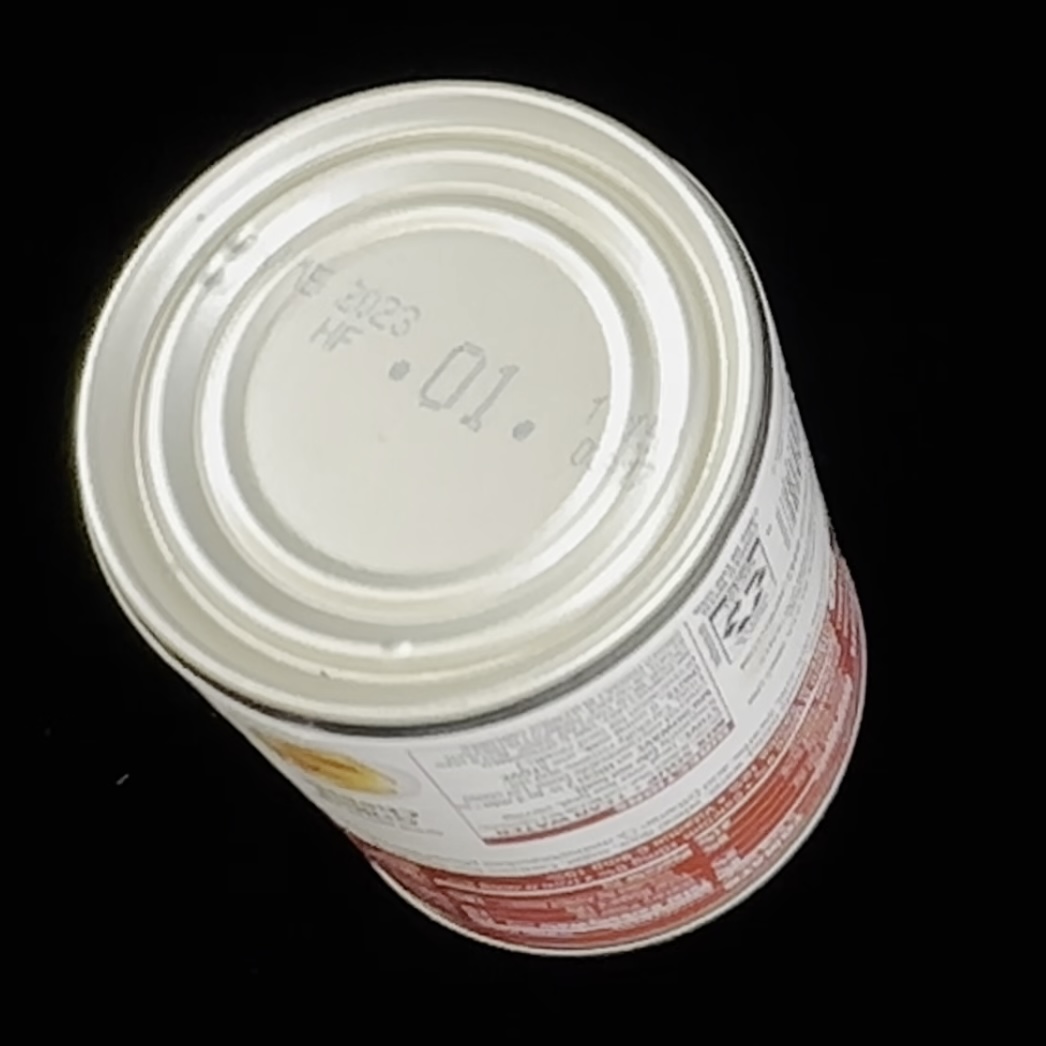}
    \end{minipage}
    \begin{minipage}{0.235\textwidth}
        \centering
        \includegraphics[width=\linewidth]{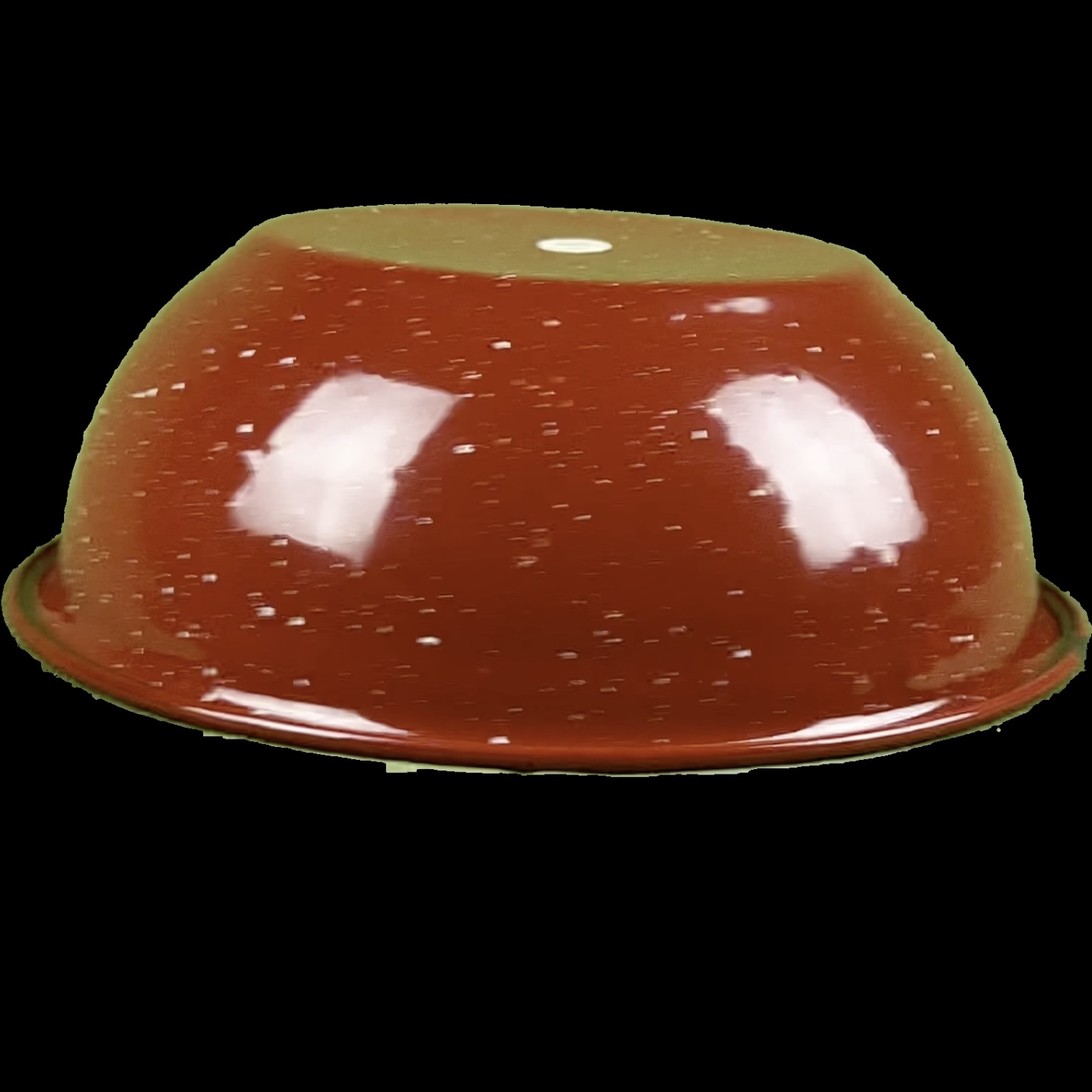}
    \end{minipage}
    \hfill
    \begin{minipage}{0.235\textwidth}
        \centering
        \includegraphics[width=\linewidth]{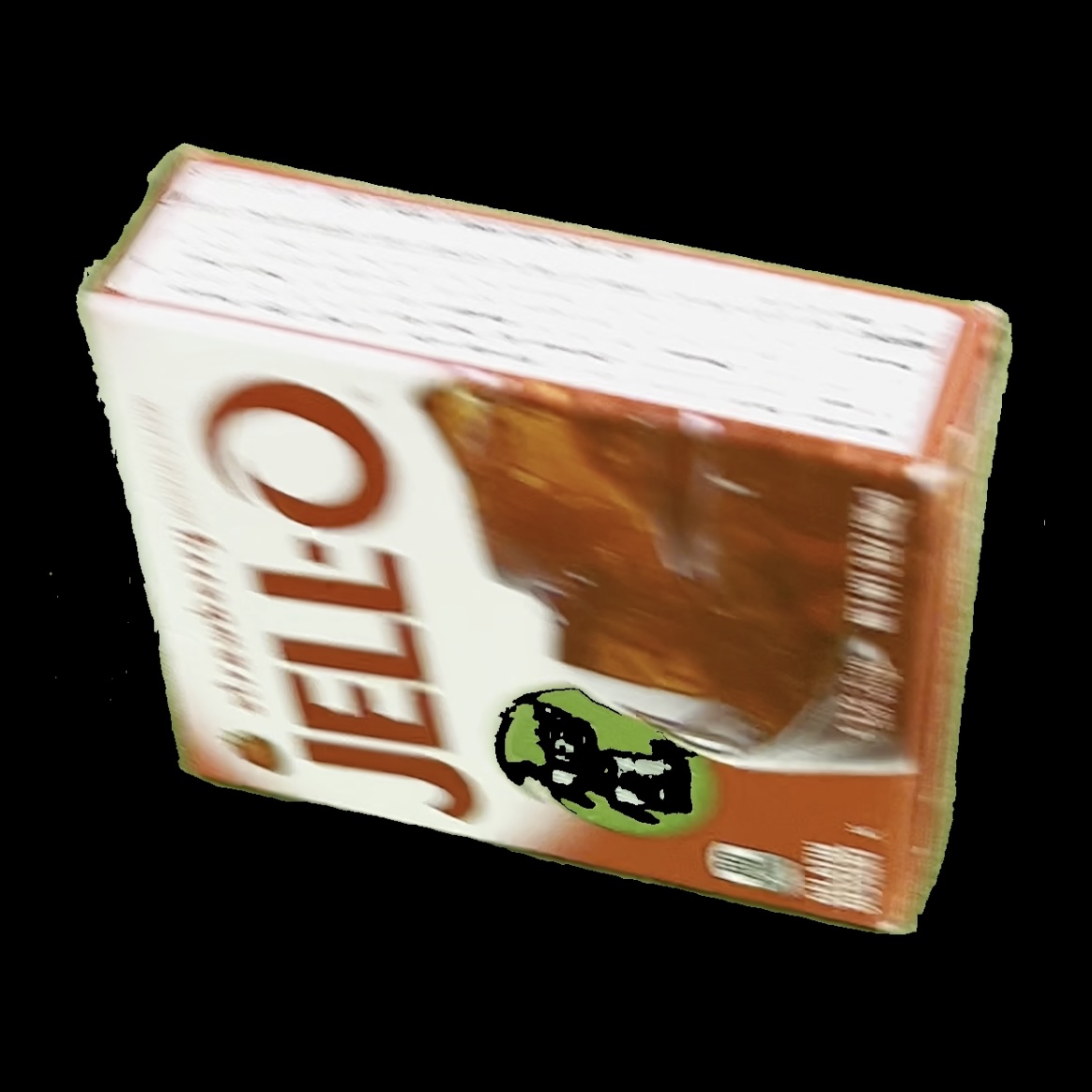}
    \end{minipage}
    \caption{Some of the problems with chroma keying. Color bleeding leads to part of the object appearing greenish, which leads to imperfect masking (top left). Luminance keying (top right) in contrast gives much improved masking. Other problems with chroma are the high reflectivity of conventional backgrounds, that lead to a "halo" effect at the edges (bottom left), and the cutting out of object parts close to the background color (bottom right).}
    \label{fig:qualitative}
\end{figure}

\subsection{Qualitative Results}
\label{sec:qualitative}
% We train a network for qualitative demonstration of our prosoed method. We generate a dataset with a resolution of 1024 x 1024 and train YOLOX with configuration X. For evaluation we recorded multiple scenes and run inference with our model.
% \textcolor{red}{TODO: show stills from the recors + link to videos + description of what we see}

\begin{figure}[htbp]
    \centering
    \begin{minipage}{0.235\textwidth}
        \centering
        \includegraphics[width=\linewidth]{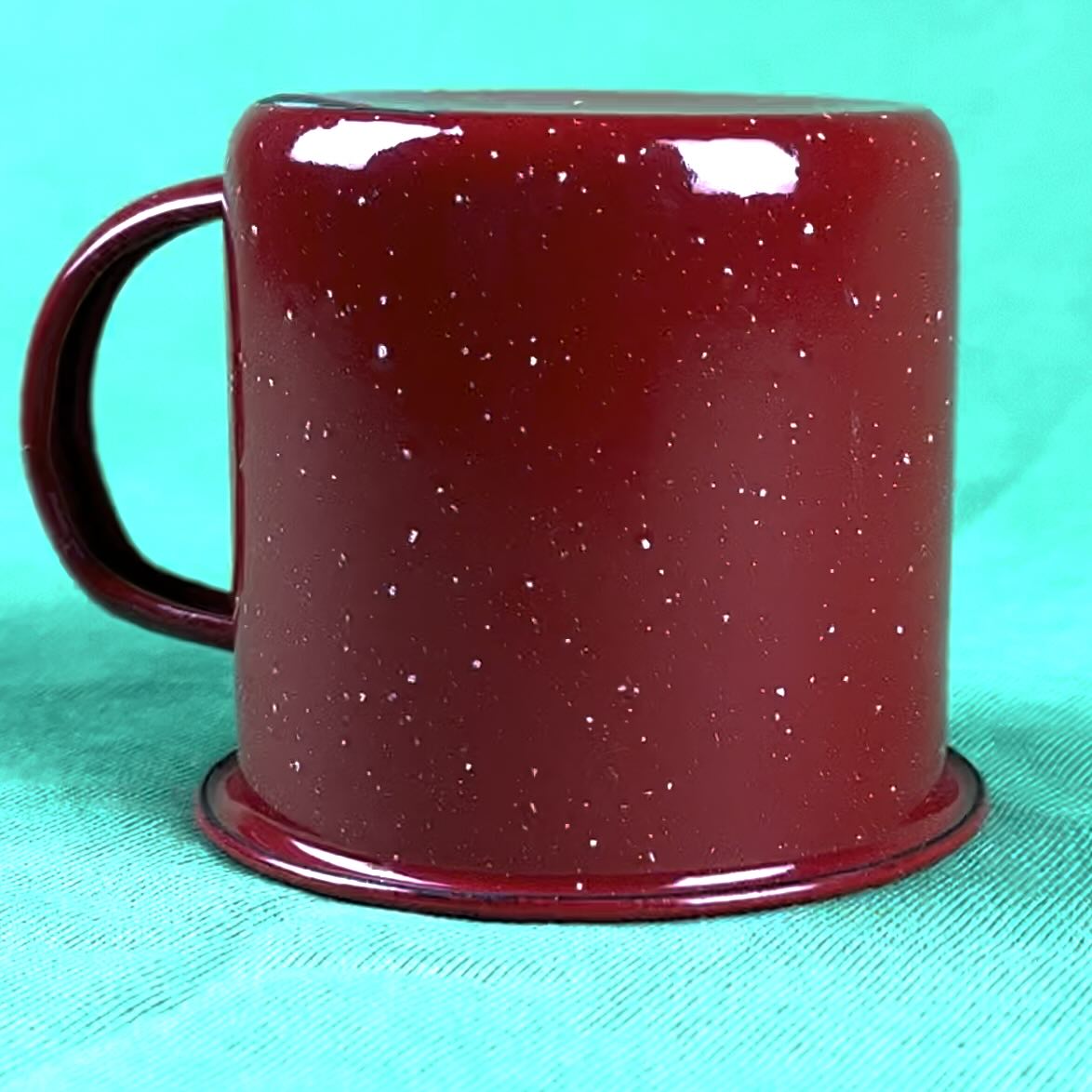}
    \end{minipage}
    \hfill
    \begin{minipage}{0.235\textwidth}
        \centering
        \includegraphics[width=\linewidth]{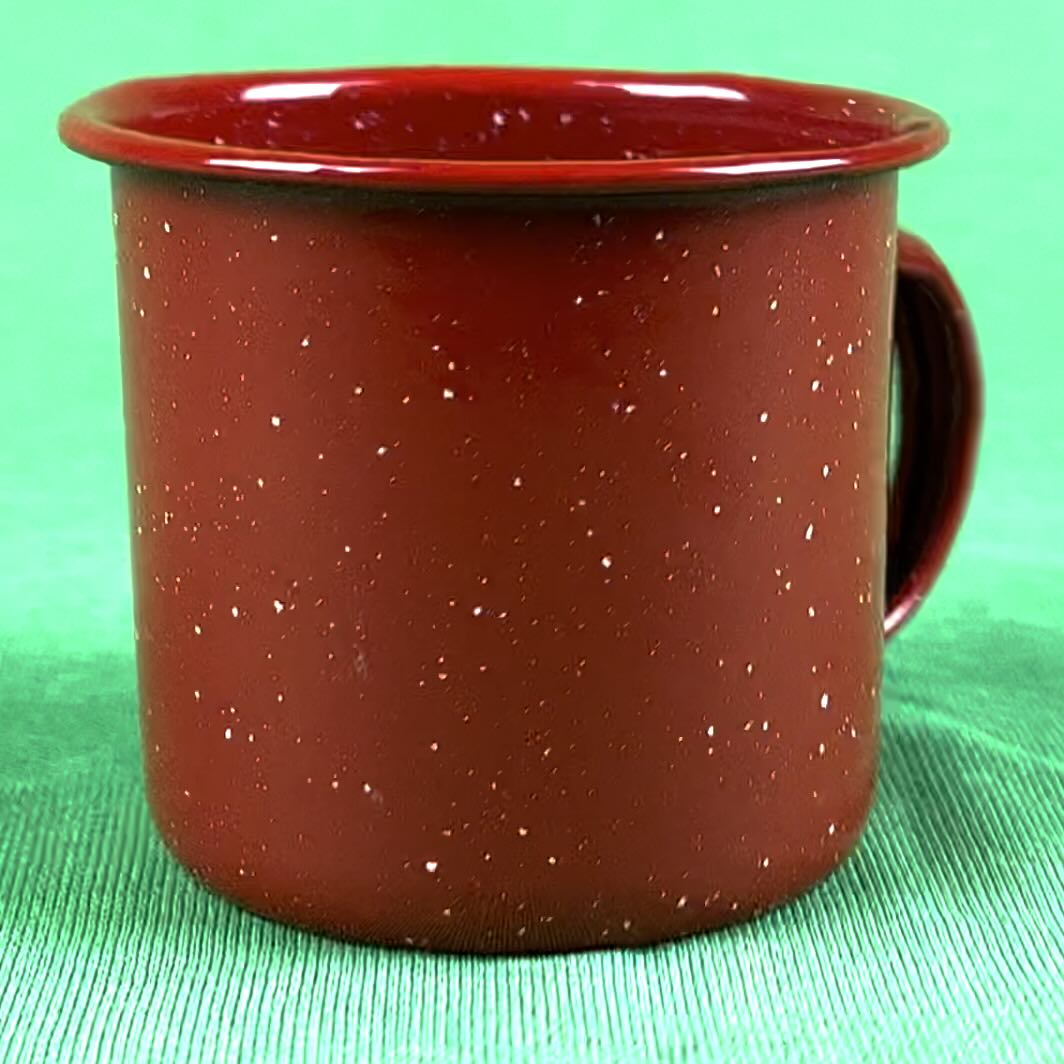}
    \end{minipage}
    \caption{Another problem of chroma keying. While the lighting did not change, a slight change in camera settings between two video clips lead to very different tones of green, making the thresholding much harder compared to luminance keying.}
    \label{fig:color_matching}
\end{figure}

We depict some of the problems of chroma keying in Figure \ref{fig:qualitative} and show the quality of our approach in Figure \ref{fig:ycbv_objects}. In direct comparison with chroma keying using the green screen we note a significant improvement in mask quality. At the same time the process becomes noticeably faster and less error-prone since we do not have to match a specific color when thresholding to generate our masks. This problem is illustrated in Figure \ref{fig:color_matching}.

\subsection{Discussion}
We see a noticable drop in performance related to our background replacement. This is to be expected since we remove relevant cues for the 3D scene understanding, such as global illumination, shadows, color bleeding, and realistic occlusion. However, when considering the subset of data representations utilizing the background replacement, namely REAL, PBR, CHROMA, and LUMA, we see the overperformance of LUMA. At the same time this relative gain in performance of REAL and PBR comes at a cost: creating a suitable big dataset to be able to train on REAL requires (usually manual) labeling, while the creation of a PBR dataset presumes the existence of 3D meshes or requires the scanning of the objects. Also, we have a throughput of roughly 50 images rendered per minute on a single A100 GPU, making the PBR dataset costly in terms of resources and time.

\section{Conclusion}
In conclusion, we propose a luminance keying method using a black screen with 99.99\% light absorption for efficient training data acquisition, significantly simplifying the process of training deep neural networks for object segmentation and detection. Our technique overcomes the limitations of manual annotation and rendering, traditionally required for creating annotated datasets, by employing a high-absorption black background to facilitate quick video recording and brightness-based automatic masking of objects. This approach not only expedites the data preparation process but also eliminates common issues associated with chroma keying, such as color bleeding and color overlap. We did extensive evaluation and find that our method compares favourably to much more involved data generation approaches on the YCB-V object set. Overall, this work enables the rapid deployment of deep learning applications across various scales, democratizing access to state-of-the-art object detection and segmentation technologies. For reproducability and further research we publish our processing code, as well as our black screen YCB-V recordings at \url{https://huggingface.co/datasets/tpoellabauer/YCB-V-LUMA}.

% \section{Future Work}
% As we have seen, we loose a substantial amount of performance when applying our background replacement script. Though we nevertheless achieve satisfactory performance, improving the processing method should further reduce the gap to using real-world recordings or physically-based renderings. In addition we plan to illustrate the superiority of luminance versus chroma keying on objects with challenging characteristics, such as transparent and metallic objects.

\bibliographystyle{alpha}
\bibliography{bib}

\newcommand{\etalchar}[1]{$^{#1}$}
\begin{thebibliography}{KKvB{\etalchar{+}}22}

\bibitem[AAPS16]{aksoy2016interactive}
Ya{\u{g}}iz Aksoy, Tun{\c{c}}~Ozan Aydin, Marc Pollefeys, and Aljo{\v{s}}a
  Smoli{\'c}.
\newblock Interactive high-quality green-screen keying via color unmixing.
\newblock {\em ACM Transactions on Graphics (TOG)}, 36(4):1, 2016.

\bibitem[AYK07]{agata2007chroma}
Hiroki Agata, Atsushi Yamashita, and Toru Kaneko.
\newblock Chroma key using a checker pattern background.
\newblock {\em IEICE TRANSACTIONS on Information and Systems}, 90(1):242--249,
  2007.

\bibitem[BIRS22]{bessen2022role}
James Bessen, Stephen~Michael Impink, Lydia Reichensperger, and Robert Seamans.
\newblock The role of data for ai startup growth.
\newblock {\em Research Policy}, 51(5):104513, 2022.

\bibitem[BRS{\etalchar{+}}22]{block2022image}
Lukas Block, Adrian Raiser, Lena Sch{\"o}n, Franziska Braun, and Oliver Riedel.
\newblock Image-bot: generating synthetic object detection datasets for small
  and medium-sized manufacturing companies.
\newblock {\em Procedia CIRP}, 107:434--439, 2022.

\bibitem[DBGD24]{cutPaste}
Jonas Dirr, Johannes~C Bauer, Daniel Gebauer, and R{\"u}diger Daub.
\newblock Cut-paste image generation for instance segmentation for robotic
  picking of industrial parts.
\newblock {\em The International Journal of Advanced Manufacturing Technology},
  130(1):191--201, 2024.

\bibitem[DMH17]{cutPasteLearn}
Debidatta Dwibedi, Ishan Misra, and Martial Hebert.
\newblock Cut, paste and learn: Surprisingly easy synthesis for instance
  detection.
\newblock In {\em Proceedings of the IEEE international conference on computer
  vision}, pages 1301--1310, 2017.

\bibitem[DSW{\etalchar{+}}20]{blenderproc}
Maximilian Denninger, Martin Sundermeyer, Dominik Winkelbauer, Dmitry Olefir,
  Tomas Hodan, Youssef Zidan, Mohamad Elbadrawy, Markus Knauer, Harinandan
  Katam, and Ahsan Lodhi.
\newblock Blenderproc: Reducing the reality gap with photorealistic rendering.
\newblock In {\em International Conference on Robotics: Sciene and Systems, RSS
  2020}, 2020.

\bibitem[Fos10]{bookgreen}
Jeff Foster.
\newblock {\em The Green Screen Handbook}.
\newblock Sybex (Wiley), 2010.

\bibitem[GKTB10]{grundhofer2010color}
Anselm Grundh{\"o}fer, Daniel Kurz, Sebastian Thiele, and Oliver Bimber.
\newblock Color invariant chroma keying and color spill neutralization for
  dynamic scenes and cameras.
\newblock {\em The Visual Computer}, 26:1167--1176, 2010.

\bibitem[GLW{\etalchar{+}}21]{yolox}
Zheng Ge, Songtao Liu, Feng Wang, Zeming Li, and Jian Sun.
\newblock Yolox: Exceeding yolo series in 2021.
\newblock {\em arXiv preprint arXiv:2107.08430}, 2021.

\bibitem[HJS{\etalchar{+}}20]{hamon2020robustness}
Ronan Hamon, Henrik Junklewitz, Ignacio Sanchez, et~al.
\newblock Robustness and explainability of artificial intelligence.
\newblock {\em Publications Office of the European Union}, 207, 2020.

\bibitem[HLWK18]{hinterstoisser_freezing}
Stefan Hinterstoisser, Vincent Lepetit, Paul Wohlhart, and Kurt Konolige.
\newblock On pre-trained image features and synthetic images for deep learning.
\newblock In {\em Proceedings of the European Conference on Computer Vision
  (ECCV) Workshops}, pages 0--0, 2018.

\bibitem[HPH{\etalchar{+}}19]{hinterstoisser2019annotation}
Stefan Hinterstoisser, Olivier Pauly, Hauke Heibel, Marek Martina, and Martin
  Bokeloh.
\newblock An annotation saved is an annotation earned: Using fully synthetic
  training for object detection.
\newblock In {\em Proceedings of the IEEE/CVF international conference on
  computer vision workshops}, pages 0--0, 2019.

\bibitem[JLZ{\etalchar{+}}22]{jin2022automatic}
Yue Jin, Zhaoxin Li, Dengming Zhu, Min Shi, and Zhaoqi Wang.
\newblock Automatic and real-time green screen keying.
\newblock {\em The Visual Computer}, 38(9):3135--3147, 2022.

\bibitem[KKvB{\etalchar{+}}22]{knauthe2022alignment}
Volker Knauthe, Maurice Kraus, Max von Buelow, Tristan Wirth, Arne Rak, Laurenz
  Merth, Alexander Erbe, Christian Kontermann, Stefan Guthe, Arjan Kuijper,
  et~al.
\newblock Alignment and reassembly of broken specimens for creep ductility
  measurements.
\newblock In {\em VMV}, pages 33--40, 2022.

\bibitem[KMR{\etalchar{+}}23]{kirillov2023segment}
Alexander Kirillov, Eric Mintun, Nikhila Ravi, Hanzi Mao, Chloe Rolland, Laura
  Gustafson, Tete Xiao, Spencer Whitehead, Alexander~C Berg, Wan-Yen Lo, et~al.
\newblock Segment anything.
\newblock In {\em Proceedings of the IEEE/CVF International Conference on
  Computer Vision}, pages 4015--4026, 2023.

\bibitem[LGL{\etalchar{+}}23]{wonder3d}
Xiaoxiao Long, Yuan-Chen Guo, Cheng Lin, Yuan Liu, Zhiyang Dou, Lingjie Liu,
  Yuexin Ma, Song-Hai Zhang, Marc Habermann, Christian Theobalt, et~al.
\newblock Wonder3d: Single image to 3d using cross-domain diffusion.
\newblock {\em arXiv preprint arXiv:2310.15008}, 2023.

\bibitem[LHB04]{lecun2004learning}
Yann LeCun, Fu~Jie Huang, and Leon Bottou.
\newblock Learning methods for generic object recognition with invariance to
  pose and lighting.
\newblock In {\em Proceedings of the 2004 IEEE Computer Society Conference on
  Computer Vision and Pattern Recognition, 2004. CVPR 2004.}, volume~2, pages
  II--104. IEEE, 2004.

\bibitem[LLSH23]{blip2}
Junnan Li, Dongxu Li, Silvio Savarese, and Steven Hoi.
\newblock Blip-2: Bootstrapping language-image pre-training with frozen image
  encoders and large language models.
\newblock In {\em International conference on machine learning}, pages
  19730--19742. PMLR, 2023.

\bibitem[LTH{\etalchar{+}}22]{liang2022advances}
Weixin Liang, Girmaw~Abebe Tadesse, Daniel Ho, Li~Fei-Fei, Matei Zaharia,
  Ce~Zhang, and James Zou.
\newblock Advances, challenges and opportunities in creating data for
  trustworthy ai.
\newblock {\em Nature Machine Intelligence}, 4(8):669--677, 2022.

\bibitem[NGP{\etalchar{+}}23]{cnos}
Van~Nguyen Nguyen, Thibault Groueix, Georgy Ponimatkin, Vincent Lepetit, and
  Tomas Hodan.
\newblock Cnos: A strong baseline for cad-based novel object segmentation.
\newblock In {\em Proceedings of the IEEE/CVF International Conference on
  Computer Vision}, pages 2134--2140, 2023.

\bibitem[ODM{\etalchar{+}}23]{dinov2}
Maxime Oquab, Timoth{\'e}e Darcet, Th{\'e}o Moutakanni, Huy Vo, Marc
  Szafraniec, Vasil Khalidov, Pierre Fernandez, Daniel Haziza, Francisco Massa,
  Alaaeldin El-Nouby, et~al.
\newblock Dinov2: Learning robust visual features without supervision.
\newblock {\em arXiv preprint arXiv:2304.07193}, 2023.

\bibitem[PJS17]{phoka2017fine}
Thanathorn Phoka, Warayu Jariyawattanarat, and Attawith Sudsang.
\newblock Fine tuning for green screen matting.
\newblock In {\em 2017 9th International Conference on Knowledge and Smart
  Technology (KST)}, pages 317--322. IEEE, 2017.

\bibitem[Ram15]{bookexperimental}
Kathryn Ramey.
\newblock {\em Experimental Filmmaking : Break the Machine}.
\newblock Taylor \& Francis Ltd, 2015.

\bibitem[RHW19]{roh2019survey}
Yuji Roh, Geon Heo, and Steven~Euijong Whang.
\newblock A survey on data collection for machine learning: a big data-ai
  integration perspective.
\newblock {\em IEEE Transactions on Knowledge and Data Engineering},
  33(4):1328--1347, 2019.

\bibitem[RKH{\etalchar{+}}21]{clip}
Alec Radford, Jong~Wook Kim, Chris Hallacy, Aditya Ramesh, Gabriel Goh,
  Sandhini Agarwal, Girish Sastry, Amanda Askell, Pamela Mishkin, Jack Clark,
  et~al.
\newblock Learning transferable visual models from natural language
  supervision.
\newblock In {\em International conference on machine learning}, pages
  8748--8763. PMLR, 2021.

\bibitem[SB96]{smith1996blue}
Alvy~Ray Smith and James~F Blinn.
\newblock Blue screen matting.
\newblock In {\em Proceedings of the 23rd annual conference on Computer
  graphics and interactive techniques}, pages 259--268, 1996.

\bibitem[Sch05]{Photogrammetry_introduction}
Toni Schenk.
\newblock Introduction to photogrammetry.
\newblock {\em The Ohio State University, Columbus}, 106(1), 2005.

\bibitem[SHL{\etalchar{+}}23]{bop22}
Martin Sundermeyer, Tom{\'a}{\v{s}} Hoda{\v{n}}, Yann Labbe, Gu~Wang, Eric
  Brachmann, Bertram Drost, Carsten Rother, and Ji{\v{r}}{\'\i} Matas.
\newblock Bop challenge 2022 on detection, segmentation and pose estimation of
  specific rigid objects.
\newblock In {\em Proceedings of the IEEE/CVF Conference on Computer Vision and
  Pattern Recognition}, pages 2784--2793, 2023.

\bibitem[TFR{\etalchar{+}}17]{Domain_randomization}
Josh Tobin, Rachel Fong, Alex Ray, Jonas Schneider, Wojciech Zaremba, and
  Pieter Abbeel.
\newblock Domain randomization for transferring deep neural networks from
  simulation to the real world.
\newblock In {\em 2017 IEEE/RSJ International Conference on Intelligent Robots
  and Systems (IROS)}, pages 23--30, 2017.

\bibitem[TPA{\etalchar{+}}18]{Bridging}
Jonathan Tremblay, Aayush Prakash, David Acuna, Mark Brophy, Varun Jampani, Cem
  Anil, Thang To, Eric Cameracci, Shaad Boochoon, and Stan Birchfield.
\newblock Training deep networks with synthetic data: Bridging the reality gap
  by domain randomization.
\newblock In {\em Proceedings of the IEEE Conference on Computer Vision and
  Pattern Recognition (CVPR) Workshops}, June 2018.

\bibitem[TTS{\etalchar{+}}18]{dope}
Jonathan Tremblay, Thang To, Balakumar Sundaralingam, Yu~Xiang, Dieter Fox, and
  Stan Birchfield.
\newblock Deep object pose estimation for semantic robotic grasping of
  household objects.
\newblock {\em arXiv preprint arXiv:1809.10790}, 2018.

\bibitem[VYB{\etalchar{+}}24]{sv3d}
Vikram Voleti, Chun-Han Yao, Mark Boss, Adam Letts, David Pankratz, Dmitry
  Tochilkin, Christian Laforte, Robin Rombach, and Varun Jampani.
\newblock Sv3d: Novel multi-view synthesis and 3d generation from a single
  image using latent video diffusion.
\newblock {\em arXiv preprint arXiv:2403.12008}, 2024.

\bibitem[WGS{\etalchar{+}}21]{distributionshift}
Olivia Wiles, Sven Gowal, Florian Stimberg, Sylvestre-Alvise Rebuffi, Ira
  Ktena, Krishnamurthy~Dj Dvijotham, and Ali~Taylan Cemgil.
\newblock A fine-grained analysis on distribution shift.
\newblock In {\em International Conference on Learning Representations}, 2021.

\bibitem[WRSL23]{whang2023data}
Steven~Euijong Whang, Yuji Roh, Hwanjun Song, and Jae-Gil Lee.
\newblock Data collection and quality challenges in deep learning: A
  data-centric ai perspective.
\newblock {\em The VLDB Journal}, 32(4):791--813, 2023.

\bibitem[YAK08]{yamashita2008every}
Atsushi Yamashita, Hiroki Agata, and Toru Kaneko.
\newblock Every color chromakey.
\newblock In {\em 2008 19th International Conference on Pattern Recognition},
  pages 1--4. IEEE, 2008.

\bibitem[YCT{\etalchar{+}}23]{hqImageDataset}
Qinhong Yang, Dongdong Chen, Zhentao Tan, Qiankun Liu, Qi~Chu, Jianmin Bao,
  Lu~Yuan, Gang Hua, and Nenghai Yu.
\newblock Hq-50k: A large-scale, high-quality dataset for image restoration.
\newblock {\em arXiv preprint arXiv:2306.05390}, 2023.

\end{thebibliography}
\
\subsection*{Appendix}

% Table generated by Excel2LaTeX from sheet 'Tiny (formatted)'
\begin{sidewaystable*}[htbp]
    \setlength{\tabcolsep}{0.5pt}
  \small
  \caption{In our extensive evaluation we address the problem of domain gap / distribution shift between image sources. First we analyse the effect of freezing the backbone (first table), training from scratch (second table), and continuing training from weights pre-trained on COCO (third table). Next we train on our 7 proposed data representations. To emphasize the influence of data similarity, we next use a range of different validation sets, corresponding to different data representations and sources and determine the best checkpoint for each. Finally, we test the best checkpoint against relevant test sets. All sets and splits are disjoint. NoBGC .. Chroma data without background replacement, NoBGL .. Luma w/o bg. repl., 50 .. a set consisting of 50 percent images with and 50\% images w/o bg. repl.}
    \begin{tabular}{|c|c|cccc|cccc|cccc|cccc|cccc|ccccccc|ccccc|cc|}
    \toprule
    \multicolumn{36}{|c|}{BB Freezing, Pretrained on COCO} \\
    \midrule
    Train &       & \multicolumn{4}{c|}{PBR rand. Tex} & \multicolumn{4}{c|}{PBR}      & \multicolumn{4}{c|}{PBG}      & \multicolumn{4}{c|}{Real}     & \multicolumn{4}{c|}{RBG}      & \multicolumn{5}{c}{Chroma}            &       &       & \multicolumn{7}{c|}{Luma} \\
    Val   &       & \multicolumn{1}{c}{PBR} & \multicolumn{1}{c}{PBG} & \multicolumn{1}{c}{Real} & \multicolumn{1}{c|}{RBG} & \multicolumn{1}{c}{PBR} & \multicolumn{1}{c}{PBG} & \multicolumn{1}{c}{Real} & \multicolumn{1}{c|}{RBG} & \multicolumn{1}{c}{PBR} & \multicolumn{1}{c}{PBG} & \multicolumn{1}{c}{Real} & \multicolumn{1}{c|}{RBG} & \multicolumn{1}{c}{PBR} & \multicolumn{1}{c}{PBG} & \multicolumn{1}{c}{Real} & \multicolumn{1}{c|}{RBG} & \multicolumn{1}{c}{PBR} & \multicolumn{1}{c}{PBG} & \multicolumn{1}{c}{Real} & \multicolumn{1}{c|}{RBG} & \multicolumn{1}{c}{PBR} & \multicolumn{1}{c}{PBG} & \multicolumn{1}{c}{Real} & \multicolumn{1}{c}{RBG} & \multicolumn{1}{c}{Chroma} & \multicolumn{1}{c}{50} & \multicolumn{1}{c|}{noBG} & \multicolumn{1}{c}{PBR} & \multicolumn{1}{c}{PBG} & \multicolumn{1}{c}{Real} & \multicolumn{1}{c}{RBG} & \multicolumn{1}{c}{Luma} & \multicolumn{1}{c}{50} & \multicolumn{1}{c|}{noBG} \\
    \multirow{6}[0]{*}{\begin{sideways}        Test\end{sideways}} & PBR   & 1.0   & 0.6   & 1.0   & 1.0   & \multicolumn{1}{c}{35.8} & \multicolumn{1}{c}{35.8} & \multicolumn{1}{c}{35.7} & \multicolumn{1}{c|}{35.7} & \multicolumn{1}{c}{9.9} & \multicolumn{1}{c}{9.8} & \multicolumn{1}{c}{9.2} & \multicolumn{1}{c|}{9.8} & \multicolumn{1}{c}{6.8} & \multicolumn{1}{c}{6.4} & \multicolumn{1}{c}{6.3} & \multicolumn{1}{c|}{6.4} & \multicolumn{1}{c}{5.8} & \multicolumn{1}{c}{5.7} & \multicolumn{1}{c}{5.7} & \multicolumn{1}{c|}{5.7} & \multicolumn{1}{c}{6.2} & \multicolumn{1}{c}{6.1} & \multicolumn{1}{c}{6.2} & \multicolumn{1}{c}{6.1} & \multicolumn{1}{c}{6.1} & \multicolumn{1}{c}{6.1} & \multicolumn{1}{c|}{6.1} & \multicolumn{1}{c}{8.1} & \multicolumn{1}{c}{8.1} & \multicolumn{1}{c}{8.1} & \multicolumn{1}{c}{8.1} & \multicolumn{1}{c}{8.1} & \multicolumn{1}{c}{8.1} & \multicolumn{1}{c|}{7.6} \\
          & PBG   & 0.2   & 0.2   & 0.2   & 0.2   & \multicolumn{1}{c}{4.6} & \multicolumn{1}{c}{4.6} & \multicolumn{1}{c}{4.6} & \multicolumn{1}{c|}{4.6} & \multicolumn{1}{c}{10.8} & \multicolumn{1}{c}{11.0} & \multicolumn{1}{c}{10.2} & \multicolumn{1}{c|}{10.9} & \multicolumn{1}{c}{2.1} & \multicolumn{1}{c}{2.1} & \multicolumn{1}{c}{2.1} & \multicolumn{1}{c|}{2.1} & \multicolumn{1}{c}{6.5} & \multicolumn{1}{c}{6.6} & \multicolumn{1}{c}{6.6} & \multicolumn{1}{c|}{6.6} & \multicolumn{1}{c}{4.9} & \multicolumn{1}{c}{4.9} & \multicolumn{1}{c}{4.9} & \multicolumn{1}{c}{5.0} & \multicolumn{1}{c}{5.0} & \multicolumn{1}{c}{5.0} & \multicolumn{1}{c|}{5.0} & \multicolumn{1}{c}{5.5} & \multicolumn{1}{c}{5.5} & \multicolumn{1}{c}{5.6} & \multicolumn{1}{c}{5.6} & \multicolumn{1}{c}{5.5} & \multicolumn{1}{c}{5.6} & \multicolumn{1}{c|}{5.3} \\
          & Real  & 0.8   & 0.2   & 0.9   & 0.8   & \multicolumn{1}{c}{43.3} & \multicolumn{1}{c}{43.3} & \multicolumn{1}{c}{43.2} & \multicolumn{1}{c|}{43.1} & \multicolumn{1}{c}{22.1} & \multicolumn{1}{c}{21.1} & \multicolumn{1}{c}{22.5} & \multicolumn{1}{c|}{21.1} & \multicolumn{1}{c}{48.1} & \multicolumn{1}{c}{50.7} & \multicolumn{1}{c}{50.9} & \multicolumn{1}{c|}{50.7} & \multicolumn{1}{c}{26.2} & \multicolumn{1}{c}{25.6} & \multicolumn{1}{c}{25.6} & \multicolumn{1}{c|}{25.3} & \multicolumn{1}{c}{21.7} & \multicolumn{1}{c}{21.4} & \multicolumn{1}{c}{21.7} & \multicolumn{1}{c}{21.3} & \multicolumn{1}{c}{21.3} & \multicolumn{1}{c}{21.3} & \multicolumn{1}{c|}{21.8} & \multicolumn{1}{c}{24.6} & \multicolumn{1}{c}{24.2} & \multicolumn{1}{c}{24.3} & \multicolumn{1}{c}{24.3} & \multicolumn{1}{c}{24.3} & \multicolumn{1}{c}{24.1} & \multicolumn{1}{c|}{23.9} \\
          & RBG   & 0.0   & 0.0   & 0.0   & 0.0   & \multicolumn{1}{c}{7.1} & \multicolumn{1}{c}{7.1} & \multicolumn{1}{c}{7.3} & \multicolumn{1}{c|}{7.0} & \multicolumn{1}{c}{15.5} & \multicolumn{1}{c}{15.3} & \multicolumn{1}{c}{15.6} & \multicolumn{1}{c|}{16.2} & \multicolumn{1}{c}{4.4} & \multicolumn{1}{c}{6.3} & \multicolumn{1}{c}{6.1} & \multicolumn{1}{c|}{6.3} & \multicolumn{1}{c}{20.7} & \multicolumn{1}{c}{21.2} & \multicolumn{1}{c}{21.1} & \multicolumn{1}{c|}{21.0} & \multicolumn{1}{c}{13.4} & \multicolumn{1}{c}{12.9} & \multicolumn{1}{c}{13.6} & \multicolumn{1}{c}{13.7} & \multicolumn{1}{c}{13.7} & \multicolumn{1}{c}{13.7} & \multicolumn{1}{c|}{12.3} & \multicolumn{1}{c}{16.0} & \multicolumn{1}{c}{16.3} & \multicolumn{1}{c}{16.3} & \multicolumn{1}{c}{16.3} & \multicolumn{1}{c}{16.3} & \multicolumn{1}{c}{16.3} & \multicolumn{1}{c|}{14.5} \\
          & NoBGC & 0.0   & 0.0   & 0.0   & 0.0   & \multicolumn{1}{c}{3.7} & \multicolumn{1}{c}{3.6} & \multicolumn{1}{c}{3.6} & \multicolumn{1}{c|}{3.5} & \multicolumn{1}{c}{0.1} & \multicolumn{1}{c}{0.0} & \multicolumn{1}{c}{0.1} & \multicolumn{1}{c|}{0.1} & \multicolumn{1}{c}{0.6} & \multicolumn{1}{c}{0.9} & \multicolumn{1}{c}{1.0} & \multicolumn{1}{c|}{0.8} & \multicolumn{1}{c}{0.5} & \multicolumn{1}{c}{0.6} & \multicolumn{1}{c}{0.5} & \multicolumn{1}{c|}{0.6} & \multicolumn{1}{c}{0.7} & \multicolumn{1}{c}{1.0} & \multicolumn{1}{c}{0.6} & \multicolumn{1}{c}{1.0} & \multicolumn{1}{c}{1.0} & \multicolumn{1}{c}{1.0} & \multicolumn{1}{c|}{1.4} & \multicolumn{1}{c}{0.5} & \multicolumn{1}{c}{0.9} & \multicolumn{1}{c}{0.9} & \multicolumn{1}{c}{0.9} & \multicolumn{1}{c}{0.9} & \multicolumn{1}{c}{1.0} & \multicolumn{1}{c|}{1.7} \\
          & NoBGL & 0.0   & 0.0   & 0.0   & 0.0   & \multicolumn{1}{c}{4.9} & \multicolumn{1}{c}{5.0} & \multicolumn{1}{c}{4.9} & \multicolumn{1}{c|}{5.1} & \multicolumn{1}{c}{0.1} & \multicolumn{1}{c}{0.1} & \multicolumn{1}{c}{0.1} & \multicolumn{1}{c|}{0.1} & \multicolumn{1}{c}{1.0} & \multicolumn{1}{c}{1.1} & \multicolumn{1}{c}{1.2} & \multicolumn{1}{c|}{1.1} & \multicolumn{1}{c}{0.3} & \multicolumn{1}{c}{0.4} & \multicolumn{1}{c}{0.4} & \multicolumn{1}{c|}{0.4} & \multicolumn{1}{c}{1.0} & \multicolumn{1}{c}{1.3} & \multicolumn{1}{c}{0.7} & \multicolumn{1}{c}{1.2} & \multicolumn{1}{c}{1.2} & \multicolumn{1}{c}{1.2} & \multicolumn{1}{c|}{2.1} & \multicolumn{1}{c}{1.0} & \multicolumn{1}{c}{1.5} & \multicolumn{1}{c}{1.5} & \multicolumn{1}{c}{1.5} & \multicolumn{1}{c}{1.5} & \multicolumn{1}{c}{1.7} & \multicolumn{1}{c|}{3.4} \\
    Ckpt  &       & 246   & 16    & 166   & 288   & \multicolumn{1}{c}{288} & \multicolumn{1}{c}{249} & \multicolumn{1}{c}{263} & \multicolumn{1}{c|}{246} & \multicolumn{1}{c}{190} & \multicolumn{1}{c}{289} & \multicolumn{1}{c}{95} & \multicolumn{1}{c|}{243} & \multicolumn{1}{c}{30} & \multicolumn{1}{c}{173} & \multicolumn{1}{c}{249} & \multicolumn{1}{c|}{179} & \multicolumn{1}{c}{214} & \multicolumn{1}{c}{267} & \multicolumn{1}{c}{277} & \multicolumn{1}{c|}{285} & \multicolumn{1}{c}{198} & \multicolumn{1}{c}{170} & \multicolumn{1}{c}{270} & \multicolumn{1}{c}{288} & \multicolumn{1}{c}{288} & \multicolumn{1}{c}{288} & \multicolumn{1}{c|}{94} & \multicolumn{1}{c}{200} & \multicolumn{1}{c}{248} & \multicolumn{1}{c}{246} & \multicolumn{1}{c}{246} & \multicolumn{1}{c}{288} & \multicolumn{1}{c}{249} & \multicolumn{1}{c|}{45} \\
    \multicolumn{1}{c}{} & \multicolumn{1}{r}{} &       &       &       & \multicolumn{1}{r}{} &       &       &       & \multicolumn{1}{r}{} &       &       &       & \multicolumn{1}{r}{} &       &       &       & \multicolumn{1}{r}{} &       &       &       & \multicolumn{1}{r}{} &       &       &       &       &       &       & \multicolumn{1}{r}{} &       &       &       &       & \multicolumn{1}{r}{} &       & \multicolumn{1}{r}{} \\
    \midrule
    \multicolumn{36}{|c}{Without BB Freezing, From Scratch} \\
    \midrule
    Train &       & \multicolumn{4}{c|}{PBR rand, Tex} & \multicolumn{4}{c|}{PBR}      & \multicolumn{4}{c|}{PBG}      & \multicolumn{4}{c|}{Real}     & \multicolumn{4}{c|}{RBG}      & \multicolumn{5}{c}{Chroma}            &       &       & \multicolumn{7}{c|}{Luma} \\
    Val   &       & \multicolumn{1}{c}{PBR} & \multicolumn{1}{c}{PBG} & \multicolumn{1}{c}{Real} & \multicolumn{1}{c|}{RBG} & \multicolumn{1}{c}{PBR} & \multicolumn{1}{c}{PBG} & \multicolumn{1}{c}{Real} & \multicolumn{1}{c|}{RBG} & \multicolumn{1}{c}{PBR} & \multicolumn{1}{c}{PBG} & \multicolumn{1}{c}{Real} & \multicolumn{1}{c|}{RBG} & \multicolumn{1}{c}{PBR} & \multicolumn{1}{c}{PBG} & \multicolumn{1}{c}{Real} & \multicolumn{1}{c|}{RBG} & \multicolumn{1}{c}{PBR} & \multicolumn{1}{c}{PBG} & \multicolumn{1}{c}{Real} & \multicolumn{1}{c|}{RBG} & \multicolumn{1}{c}{PBR} & \multicolumn{1}{c}{PBG} & \multicolumn{1}{c}{Real} & \multicolumn{1}{c}{RBG} & \multicolumn{1}{c}{Chroma} & \multicolumn{1}{c}{50} & \multicolumn{1}{c|}{noBG} & \multicolumn{1}{c}{PBR} & \multicolumn{1}{c}{PBG} & \multicolumn{1}{c}{Real} & \multicolumn{1}{c}{RBG} & \multicolumn{1}{c}{Luma} & \multicolumn{1}{c}{50} & \multicolumn{1}{c|}{noBG} \\
    \multirow{6}[0]{*}{\begin{sideways}        Test\end{sideways}} & PBR   & 1.8   & 0.8   & 0.1   & 1.6   & 73.7  & 72.9  & 68.7  & 72.3  & 10.4  & 2.4   & 7.6   & 6.8   & 1.6   & 1.6   & 1.6   & 1.6   & 0.7   & 0.2   & 0.5   & 0.1   & 3.5   & 3.1   & 2.9   & 3.1   & 1.6   & 1.6   & 3.1   & 6.0   & 5.1   & 5.5   & 5.1   & 2.8   & 4.2   & 5.9 \\
          & PBG   & 0.2   & 0.2   & 0.1   & 0.1   & 6.1   & 6.5   & 6.4   & 6.6   & 15.0  & 93.3  & 11.4  & 25.0  & 0.9   & 0.8   & 0.8   & 0.9   & 5.6   & 7.4   & 5.8   & 6.3   & 6.3   & 6.4   & 6.3   & 6.5   & 6.0   & 6.0   & 4.9   & 7.2   & 7.8   & 7.4   & 7.8   & 6.5   & 7.4   & 7.4 \\
          & Real  & 0.0   & 0.0   & 0.0   & 0.0   & 8.7   & 15.0  & 19.5  & 15.6  & 0.4   & 0.0   & 0.6   & 0.2   & 44.3  & 46.4  & 46.2  & 46.3  & 24.6  & 18.4  & 22.8  & 15.1  & 2.7   & 2.3   & 2.0   & 2.3   & 0.5   & 0.5   & 3.3   & 0.8   & 0.9   & 1.7   & 0.6   & 0.1   & 0.4   & 0.7 \\
          & RBG   & 0.0   & 0.0   & 0.0   & 0.0   & 0.9   & 1.0   & 2.2   & 1.2   & 4.8   & 1.3   & 1.2   & 4.7   & 2.7   & 2.6   & 1.8   & 2.1   & 35.1  & 43.3  & 36.5  & 39.0  & 8.5   & 9.6   & 7.6   & 9.2   & 4.2   & 4.2   & 6.9   & 10.0  & 10.6  & 11.0  & 10.8  & 4.0   & 9.9   & 9.6 \\
          & NoBGC & 0.0   & 0.0   & 0.0   & 0.0   & 3.5   & 2.8   & 4.2   & 3.1   & 0.0   & 0.0   & 0.1   & 0.0   & 0.5   & 0.6   & 0.5   & 0.6   & 0.0   & 0.0   & 0.0   & 0.0   & 14.2  & 13.7  & 10.3  & 12.7  & 11.0  & 11.0  & 24.2  & 3.9   & 1.0   & 5.0   & 1.2   & 1.8   & 2.2   & 5.3 \\
          & NoBGL & 0.0   & 0.0   & 0.0   & 0.0   & 7.2   & 7.4   & 8.9   & 9.8   & 0.0   & 0.0   & 0.2   & 0.0   & 0.7   & 0.9   & 1.0   & 1.2   & 0.0   & 0.0   & 0.0   & 0.0   & 4.6   & 5.1   & 4.0   & 4.5   & 6.3   & 6.3   & 6.1   & 21.8  & 16.7  & 27.5  & 23.2  & 23.8  & 28.4  & 31.5 \\
    Ckpt  &       & 48    & 6     & 2     & 101   & 169   & 80    & 43    & 63    & 18    & 289   & 6     & 39    & 103   & 161   & 182   & 165   & 29    & 165   & 32    & 271   & 48    & 63    & 85    & 64    & 250   & 250   & 22    & 54    & 188   & 75    & 172   & 288   & 219   & 60 \\
    \multicolumn{1}{c}{} & \multicolumn{1}{r}{} &       &       &       & \multicolumn{1}{r}{} &       &       &       & \multicolumn{1}{r}{} &       &       &       & \multicolumn{1}{r}{} &       &       &       & \multicolumn{1}{r}{} &       &       &       & \multicolumn{1}{r}{} &       &       &       &       &       &       & \multicolumn{1}{r}{} &       &       &       &       & \multicolumn{1}{r}{} &       & \multicolumn{1}{r}{} \\
    \midrule
    \multicolumn{36}{|c}{Without BB Freezing, Pretrained on COCO} \\
    \midrule
    Train &       & \multicolumn{4}{c|}{PBR rand, Tex} & \multicolumn{4}{c|}{PBR}      & \multicolumn{4}{c|}{PBG}      & \multicolumn{4}{c|}{Real}     & \multicolumn{4}{c|}{RBG}      & \multicolumn{5}{c}{Chroma}            &       &       & \multicolumn{7}{c|}{Luma} \\
    Val   &       & \multicolumn{1}{c}{PBR} & \multicolumn{1}{c}{PBG} & \multicolumn{1}{c}{Real} & \multicolumn{1}{c|}{RBG} & \multicolumn{1}{c}{PBR} & \multicolumn{1}{c}{PBG} & \multicolumn{1}{c}{Real} & \multicolumn{1}{c|}{RBG} & \multicolumn{1}{c}{PBR} & \multicolumn{1}{c}{PBG} & \multicolumn{1}{c}{Real} & \multicolumn{1}{c|}{RBG} & \multicolumn{1}{c}{PBR} & \multicolumn{1}{c}{PBG} & \multicolumn{1}{c}{Real} & \multicolumn{1}{c|}{RBG} & \multicolumn{1}{c}{PBR} & \multicolumn{1}{c}{PBG} & \multicolumn{1}{c}{Real} & \multicolumn{1}{c|}{RBG} & \multicolumn{1}{c}{PBR} & \multicolumn{1}{c}{PBG} & \multicolumn{1}{c}{Real} & \multicolumn{1}{c}{RBG} & \multicolumn{1}{c}{Chroma} & \multicolumn{1}{c}{50} & \multicolumn{1}{c|}{noBG} & \multicolumn{1}{c}{PBR} & \multicolumn{1}{c}{PBG} & \multicolumn{1}{c}{Real} & \multicolumn{1}{c}{RBG} & \multicolumn{1}{c}{Luma} & \multicolumn{1}{c}{50} & \multicolumn{1}{c|}{noBG} \\
    \multirow{6}[0]{*}{\begin{sideways}        Test\end{sideways}} & PBR   & 3.2   & 3.2   & 3.2   & 3.2   & 74.7  & 52.8  & 52.8  & 52.8  & 9.6   & 2.2   & 8.7   & 9.6   & 10.5  & 10.5  & 10.1  & 10.1  & 4.9   & 4.8   & 4.8   & 0.0   & 5.8   & 5.8   & 5.3   & 5.8   & 2.0   & 2.8   & 2.8   & 8.3   & 6.3   & 7.1   & 7.7   & 3.4   & 4.3   & 7.1 \\
          & PBG   & 0.6   & 0.6   & 0.6   & 0.6   & 6.2   & 6.5   & 6.5   & 6.5   & 19.8  & 94.4  & 16.6  & 19.8  & 3.3   & 3.3   & 3.4   & 3.4   & 8.9   & 9.5   & 9.5   & 7.5   & 6.6   & 6.6   & 5.4   & 6.6   & 6.3   & 6.7   & 6.7   & 6.7   & 7.2   & 5.6   & 6.4   & 6.3   & 6.7   & 5.6 \\
          & Real  & 0.5   & 0.5   & 0.5   & 0.5   & 11.4  & 26.2  & 26.2  & 26.2  & 0.0   & 0.0   & 0.1   & 0.0   & 64.2  & 64.2  & 64.3  & 64.3  & 21.8  & 24.8  & 24.8  & 14.7  & 2.6   & 2.6   & 10.0  & 2.6   & 0.5   & 1.0   & 1.0   & 2.3   & 0.5   & 7.2   & 1.7   & 0.1   & 0.1   & 7.2 \\
          & RBG   & 0.0   & 0.0   & 0.0   & 0.0   & 2.8   & 6.7   & 6.7   & 6.7   & 15.2  & 0.6   & 13.0  & 15.2  & 11.3  & 11.3  & 10.6  & 10.6  & 31.9  & 34.1  & 34.1  & 41.4  & 9.6   & 9.6   & 14.9  & 9.6   & 6.4   & 6.9   & 6.9   & 10.2  & 7.5   & 12.6  & 13.0  & 4.3   & 6.1   & 12.6 \\
          & NoBGC & 0.0   & 0.0   & 0.0   & 0.0   & 8.5   & 6.6   & 6.6   & 6.6   & 0.0   & 0.0   & 0.0   & 0.0   & 5.9   & 5.9   & 5.9   & 5.9   & 0.3   & 0.0   & 0.0   & 0.0   & 11.9  & 11.9  & 11.5  & 11.9  & 11.6  & 19.0  & 19.0  & 4.2   & 0.5   & 12.4  & 2.5   & 1.1   & 1.4   & 12.4 \\
          & NoBGL & 0.0   & 0.0   & 0.0   & 0.0   & 11.0  & 8.5   & 8.5   & 8.5   & 0.1   & 0.0   & 0.0   & 0.1   & 3.2   & 3.2   & 3.3   & 3.3   & 0.1   & 0.1   & 0.1   & 0.0   & 4.1   & 4.1   & 4.7   & 4.1   & 2.8   & 6.3   & 6.3   & 20.0  & 6.3   & 24.9  & 16.8  & 12.5  & 14.2  & 24.9 \\
    Ckpt  &       & 1     & 1     & 1     & 1     & 140   & 3     & 3     & 3     & 13    & 279   & 7     & 13    & 9     & 9     & 6     & 6     & 5     & 7     & 7     & 261   & 8     & 8     & 2     & 8     & 246   & 166   & 166   & 11    & 158   & 2     & 20    & 288   & 249   & 2 \\
    \end{tabular}%
  \label{tab:addlabel}%
\end{sidewaystable*}%

% \begin{thebibliography}{99}
% \label{references}
% \bibitem[And01a]{and01a} Anderson, R.E. Social impacts of computing: Codes of professional ethics. Social Science, pp.453-469, 2001.
% \bibitem[Con00a]{con00a} Conger., S., and Loch, K.D. (eds.). Ethics and computer use. Com.of ACM 38, No.12, 2000.
% \bibitem[Con00b]{con00b} Mackay, W.E. Ethics, lies and videotape, in Conf.proc. CHI'00, Denver CO, ACM Press, pp.138-145, 2000.
% \bibitem[Jou01a]{jou01a} Journal of WSCG \& WSCG templates: http://wscg.zcu.cz/jwscg/template.doc (MSWord)
% http://wscg.zcu.cz/jwscg/template.pdf (PDF)
% \end{thebibliography}

% {\bfseries
% Last page should be fully used by text, figures etc. Do not leave empty space, please. 

% Do not lock the PDF -- additional text and info will be inserted, i.e. ISSN/ISBN etc. 
% }

\end{document}